\documentclass[lettersize,journal]{IEEEtran}
\usepackage{amsmath,amsfonts}
\usepackage{algorithmic}
\usepackage{algorithm}
\usepackage{array}
\usepackage[caption=false,font=normalsize,labelfont=sf,textfont=sf]{subfig}
\usepackage{textcomp}
\usepackage{stfloats}
\usepackage{url}
\usepackage{verbatim}
\usepackage{graphicx}
\usepackage{cite}
\usepackage{xcolor}
\usepackage{threeparttable}
\usepackage{multirow}
\usepackage{makecell}
\usepackage{hyperref}
\usepackage{booktabs}
\usepackage{arydshln}
\usepackage{colortbl}
\usepackage[table]{xcolor}
\usepackage[capitalize]{cleveref}
\begin{document}

\title{RealCustom++: Representing Images as \\ Real Textual Word for Real-Time Customization}

\author{Zhendong Mao,~\IEEEmembership{Member,~IEEE}, 
Mengqi Huang,
Fei Ding,
Mingcong Liu,
Qian He, \\
and Yongdong Zhang,~\IEEEmembership{Fellow,~IEEE} 
\thanks{Zhendong Mao, Mengqi Huang, Yongdong Zhang are with the University of Science and Technology of China. E-mail: \{zdmao, zhyd73\}@ustc.edu.cn, huangmq@mail.ustc.edu.cn. Fei Ding, Mingcong Liu, Qian He are with the ByteDance Inc.
E-mail: \{dingfei.212, liumingcong, heqian\}@bytedance.com.} 
\thanks{This work was supported by Artificial Intelligence National Science and Technology Major Project
2023ZD0121200, and National Natural Science Foundation of China under Grant 62222212, 623B2094 and 62121002.}
\thanks{Corresponding author: Yongdong Zhang.}
}

\markboth{Journal of \LaTeX\ Class Files,~Vol.~14, No.~8, August~2021}%
{Shell \MakeLowercase{\textit{et al.}}: A Sample Article Using IEEEtran.cls for IEEE Journals}

\IEEEpubid{0000--0000/00\$00.00~\copyright~2021 IEEE}

\maketitle

\begin{abstract}
Given a text and an image of a specific subject, text-to-image customization aims to generate new images that align with both the text and the subject’s appearance. Existing works follow the pseudo-word paradigm, which represents the subject as a non-existent pseudo word and combines it with other text to generate images. However, the pseudo word causes semantic conflict from its different learning objective and entanglement from overlapping influence scopes with other texts, resulting in a dual-optimum paradox where subject similarity and text controllability cannot be optimal simultaneously. To address this, we propose RealCustom++, a novel real-word paradigm that represents the subject with a non-conflicting real word to firstly generate a coherent guidance image and corresponding subject mask, thereby disentangling the influence scopes of the text and subject for simultaneous optimization. Specifically, RealCustom++ introduces a train-inference decoupled framework: (1) during training, it learns a general alignment between visual conditions and all real words in the text; and (2) during inference, a dual-branch architecture is employed, where the Guidance Branch produces the subject guidance mask and the Generation Branch utilizes this mask to customize the generation of the specific real word exclusively within subject-relevant regions. In contrast to previous methods that excel in either controllability or similarity, RealCustom++ achieves superior performance in both, with improvements of 7.48\% in controllability, 3.04\% in similarity, and 76.43\% in generation quality. For multi-subject customization, RealCustom++ further achieves improvements of 4.6\% in controllability and 6.34\% in multi-subject similarity. Our work has been applied in JiMeng of ByteDance, and codes are released at \url{https://github.com/bytedance/RealCustom}.
\end{abstract}



\section{Introduction}

\begin{figure*}
  \centering
  \includegraphics[width=0.85\linewidth]{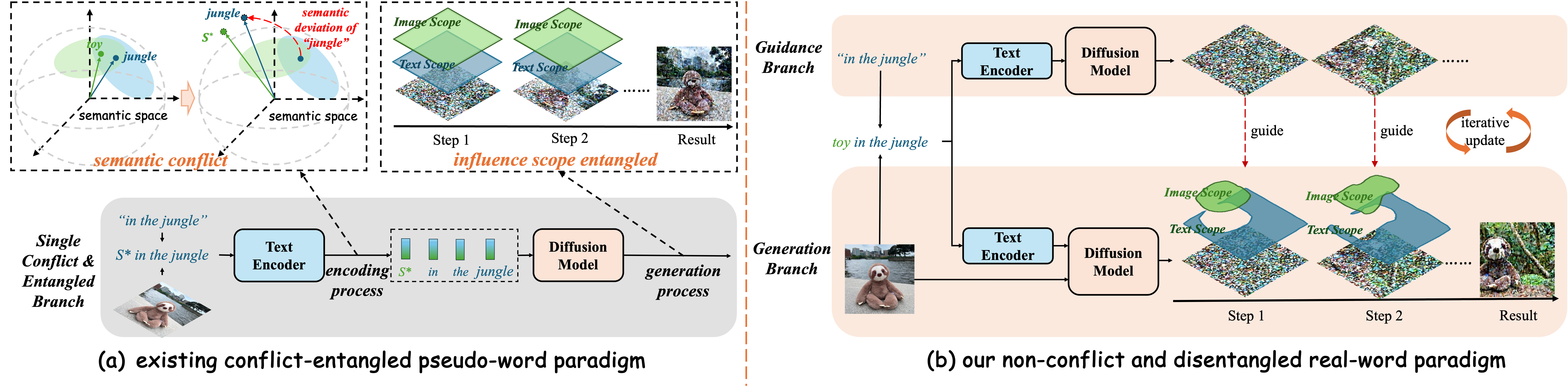}
  \caption{\textbf{(a)} Existing paradigm represents the subject as a pseudo word ($S^*$) and combines it with the text for generation. The pseudo word inherently conflicts (\emph{i.e.}, causes other real words to deviate from their original semantics) and entangles (\emph{i.e.}, has overlapping influence scope) with the text, resulting in the dual-optimum paradox that involves a trade-off between subject similarity and text controllability. \textbf{(b)} RealCustom++ first represents the subject as real words (\emph{e.g.}, the subject's super-category) to generate a guidance image in the guidance branch, providing the subject guidance mask. Then, in the generation branch, the subject influences only within the mask, while other regions are controlled purely by the text, achieving both high similarity and controllability.}
  \label{introduction}
\end{figure*}

\IEEEPARstart{T}{ext}-to-image customization~\cite{wang2024ms, han2024ace, pan2024locate, croitoru2023diffusion, gal2022image, ruiz2023dreambooth}, which takes \emph{given texts} and images of \emph{given subjects} as inputs, aims to synthesize new images that are consistent with both textual semantics and the subjects' appearance. Compared to text-to-image generation~\cite{ramesh2021zero, zhang2018stackgan++, sun2024anyface++, hinz2020semantic}, this task further offers precise control over visual details by allowing modifying any specific subjects using the text (\emph{e.g.}, making your pet wear Iron Man's suit), which is crucial for real-world applications such as film production, attracting increasing interest from both academia and industry. Moreover, this task is more challenges as its primary goals are dual-faceted: (1) \textbf{\emph{high subject-similarity}}, \emph{i.e.}, the generated subjects should closely mirror \emph{given subjects}; (2) \textbf{\emph{high text-controllability}}, \emph{i.e.}, the remaining subject-irrelevant generated parts should consistently adhere to \emph{given texts}.

\IEEEpubidadjcol

Existing customization methods follow a two-step \textbf{\emph{pseudo-word}} paradigm: (1) representing the given subject as pseudo words\cite{gal2022image, ruiz2023dreambooth}, which share the same dimensionality as real words but are non-existent in the vocabulary; (2) combining the pseudo words with other given texts to generate images. Early methods\cite{gal2022image, ruiz2023dreambooth, kumari2023multi, han2023svdiff, liu2023cones, nam2024dreammatcher, hua2023dreamtuner} individually finetune pre-trained text-to-image models\cite{rombach2022high, podell2023sdxl} for each subject, requiring minutes to hours of inference time, which limits practical applicability. Recent encoder-based methods\cite{wei2023elite, li2024blip, shi2023instantbooth, gal2023designing, li2024photomaker, wang2024instantid, guo2024pulid, xiao2023fastcomposer} learn image encoders to map subjects into pseudo words, enabling more efficient inference and therefore receiving more interest. These methods focus on developing various adapters\cite{chen2023dreamidentity, wei2023elite, li2024blip, xiao2023fastcomposer} to enhance pseudo-word influence for improved subject similarity, and different regularized losses (\emph{e.g.}, $\ell_1$ regularization\cite{wei2023elite, li2024blip, li2024photomaker}, alignment loss\cite{guo2024pulid}) to prevent overfitting caused by excessive pseudo word influence, maintaining a delicate balance. Essentially, existing methods face a dual-optimum paradox, where a trade-off exists between subject similarity and text controllability, and they cannot achieve the optimum of both simultaneously.


We argue that the dual-optimum paradox stems from the pseudo-word paradigm that combines pseudo words with given texts as a unified condition for image generation, resulting in conflicts and entanglements between the pseudo and real words. Specifically, (1) conflicts arise from their non-homologous learning objective, \emph{i.e.}, pseudo words are learned solely by reconstructing subject images and lack contextual alignment with other real words, causing semantic drift when combined (\emph{e.g.}, as shown in \cref{introduction}, the pseudo word $S^*$ distorts the semantic of the real word ``jungle", shifting it away from its original meaning and yielding non-jungle outputs). (2) Entanglements arise from their overlapping influence scopes, \emph{i.e.}, the pseudo word indiscriminately affects all generated regions, causing subject-irrelevant regions to overfit the reference image (\emph{e.g.}, as shown in \cref{introduction}, the generated background mimics the subject image rather than reflecting the intended ``in the jungle" context). \emph{A comprehensive analysis and empirical validation are provided in \cref{method_analysis_and_empirical_validation}.}

To address this problem, we present a novel paradigm, \textbf{\emph{RealCustom++}}, that, for the first time, disentangles subject similarity from text controllability, enabling both to be simultaneously optimized without conflict. Our key idea is to (1) eliminate conflict by representing the subject as real words (\emph{e.g.}, the subject's super-category), generating a coherent guidance image and corresponding subject mask, and (2) eliminate entanglement by restricting the subject’s influence only within the mask. As illustrated in \cref{introduction}(b), the sloth toy is represented as ``toy" to form the non-conflicting condition ``toy in the jungle", which first generates a text-guided image in the guidance branch to provide the guidance mask. In the generation branch, the subject’s influence is restricted to the masked region, and this iterative process progressively refines the mask, transforming regions relevant to ``toy" into the sloth toy, while other regions are generated solely based on the text, thereby achieving both high similarity and controllability.

Technically, due to the lack of annotated pairs between real words and subjects, RealCustom++ introduces an innovative train-inference decoupled framework: (1) During training, it learns a general alignment between visual conditions and all real words in the text, which is achieved by the \emph{Cross-layer Cross-scale Projector (CCP)} to extract fine-grained and robust subject representations, and the \emph{Curriculum Training Recipe (CTR)} to smoothly and effectively inject subject representations. Specifically, the CCP module dynamically fuses multi-level image features by cross-layer attention with shallow layers' features to enhance structural robustness, and multi-scale features by cross-scale attention with high-resolution features to enrich fine-grained details. The CTR adopts an ``easy-to-hard'' data curation strategy, gradually increasing subject image diversity and complexity to enable rapid convergence on basic reconstruction before tackling challenging re-contextualization, therefore enhancing alignment with textual semantics and bridging the training-inference gap. (2) During inference, we propose a dual-branch architecture connected by an \emph{Adaptive Mask Guidance (AMG)} mechanism, where the Guidance Branch produces the subject guidance mask and the Generation Branch utilizes this mask to customize the generation of the specific real word exclusively within subject-relevant regions. The guidance mask is constructed by integrating cross-attention maps of the real word, together with self-attention maps to enhance fine-grained spatial accuracy, and an ``early stop'' regularization to improve temporal stability. Moreover, we further extend it to the \emph{Multiple Adaptive Mask Guidance (M-AMG)} for multi-subject customization.

\begin{figure}
  \centering
  \includegraphics[width=0.65\linewidth]{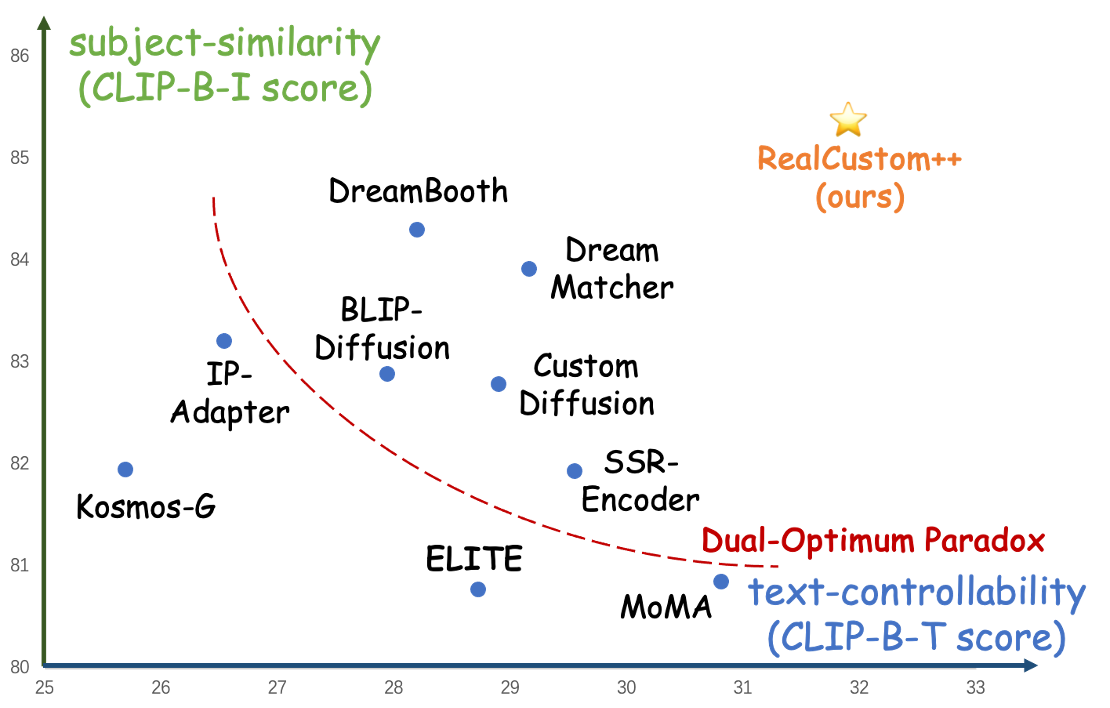}
  \caption{The quantitative comparison shows that RealCustom++ achieves the highest similarity and controllability to the existing paradigm simultaneously.}
  \label{introduction_paradox}
\end{figure}

\begin{figure*}
  \centering
  \includegraphics[width=0.9\linewidth]{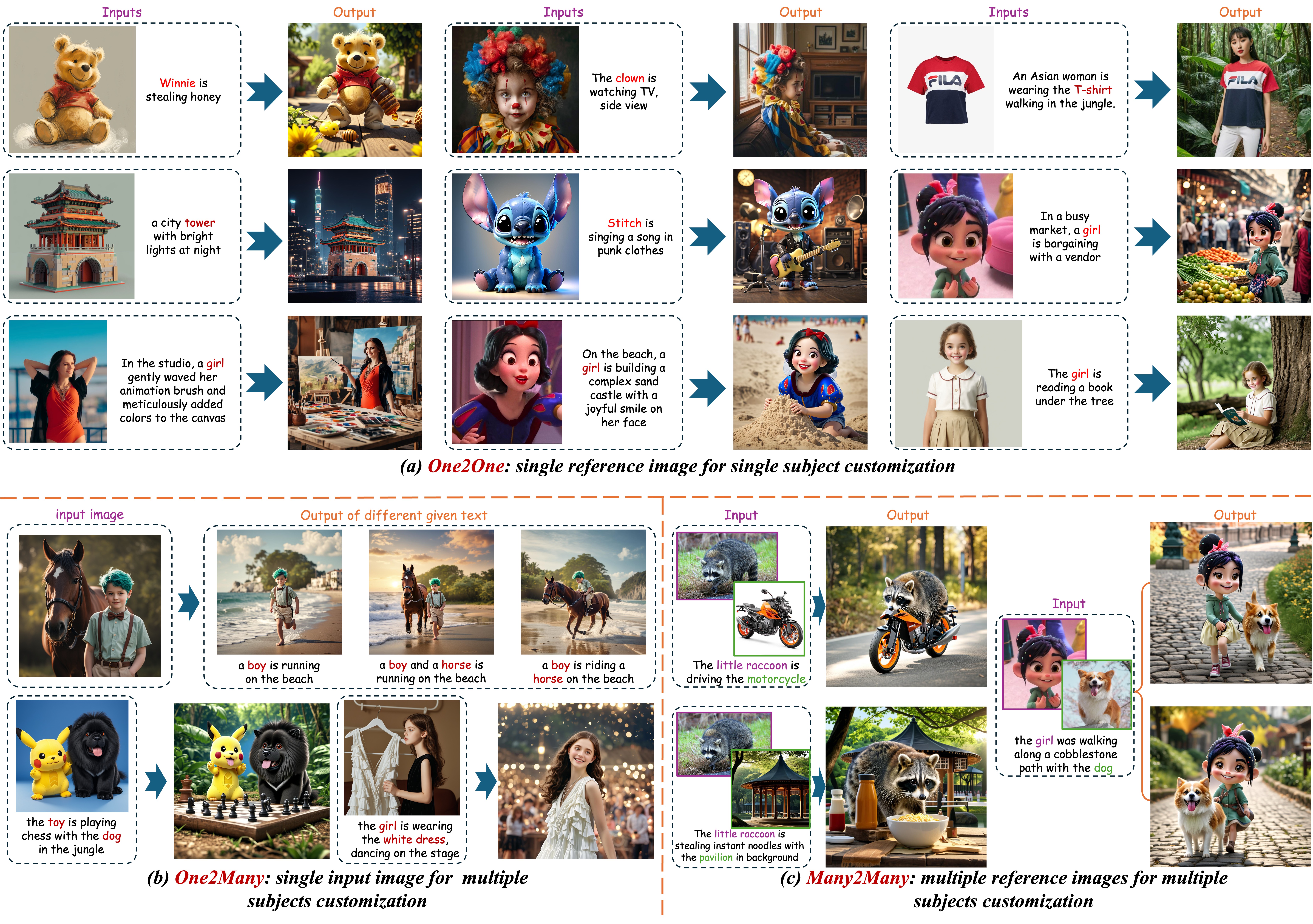}
  \caption{\textbf{\emph{Our RealCustom++ is capable of various customization tasks.}} (a) \textbf{\emph{One2One}:} Given a single image depicting the given subject (\textbf{\emph{in open domain}}, \emph{e.g.}, humans, cartoons, clothes, buildings), RealCustom++ can synthesize images that are consistent with both the semantics of the texts and the appearance of the subjects. (\textbf{\emph{in real-time without any finetuning steps}}). (b) \textbf{\emph{One2Many}}: RealCustom++ can decouple and customize each subject in a single reference image. (c) \textbf{\emph{Many2Many}}: RealCustom++ can customize multiple subjects from multiple reference images. The customized words are highlighted in color.}
  \label{introduction_cases}
\end{figure*}

As demonstrated in \cref{introduction_paradox}, RealCustom++ effectively resolves the dual-optimum paradox, achieving highest similarity and controllability simultaneously. As shown in \cref{introduction_cases}, RealCustom++ exhibits strong generalization across diverse open-domain customization tasks, owing to its ability to learn a \textbf{\emph{general alignment between visual conditions and all real words in text}} during training. This enables users to flexibly customize \textbf{\emph{any subject}} by selecting \textbf{\emph{any real word}} at inference.

We summarize our contributions as follows:

\textbf{Concept.} We identify that the dual-optimum paradox stems from the pseudo-word paradigm, and propose RealCustom++, which, for the first time, disentangles subject similarity from text controllability to enable their simultaneous optimization.

\textbf{Methodology.} We propose a novel train-inference decoupled framework that learns a general alignment between visual conditions and all real words during training, and customizes generation for the target word during inference, including a \emph{Cross-layer Cross-scale Projector (CCP)} for robust, fine-grained subject representation, a \emph{Curriculum Training Recipe (CTR)} for smooth subject injection, and an \emph{Adaptive Mask Guidance (AMG)} for subject and text disentanglement.

\textbf{Performance.} To the best of our knowledge, we are the first to simultaneously achieve highest controllability and similarity, surpassing previous state-of-the-art by 7.48\% in controllability, 3.04\% in similarity, and 76.43\% in image quality. We also improves controllability and multiple-subject similarity by 4.6\% and 6.34\% for multiple-subject customization.

\section{Related Works}

\subsection{Discussion with Conference Version}


This manuscript extends RealCustom~\cite{huang2024realcustom}, a conference paper presented at the IEEE/CVF Conference on Computer Vision and Pattern Recognition (CVPR) 2024. While sharing the same motivation, we \textbf{\emph{completely redesign our paradigm}}, as illustrated in \cref{method_discuss_conference}. Specifically, in the conference version~\cite{huang2024realcustom}, we employ a \textbf{\emph{re-construction training}} in which the reference image is identical to the generated image. To prevent the model from overfitting to the image condition, we introduce an adaptive scoring module with importance-aware feature dropping, selecting only image features most relevant to the subject. However, the dropping ratio must be \emph{manually set} to strike a similarity-controllability balance, and this reconstruction approach inherently \emph{limits the use of more fine-grained image features}, as more detailed features make the overfitting problem more severe.


To address these limitations, we propose a new \textbf{\emph{re-contextualization training paradigm}}, where the reference image exhibits varying subject sizes and poses compared to the generated image. This innovation fundamentally prevents copy-paste overfitting for higher controllability, making it possible to fine-grained feature utilization for higher similarity. To fully leverage the strengths of this new re-contextualization training paradigm, we have developed entirely new modules, such as the \emph{Curriculum Training Recipe (CTR)} that systematically controls subject size and pose variations throughout the training, the \emph{Cross-layer Cross-scale Projector (CCP)} that aggregates features from higher resolutions and multiple layers for higher similarity, \emph{etc}. We further enhance the spatial and temporal accuracy of the guidance mask to enable more precise disentanglement between similarity and controllability. Collectively, these advancements yield simultaneous improvements of 6.06\% in controllability, 2.32\% in similarity, and 56.6\% in generation quality. The major extensions include:

\begin{figure}
  \centering
  \includegraphics[width=0.9
  \linewidth]{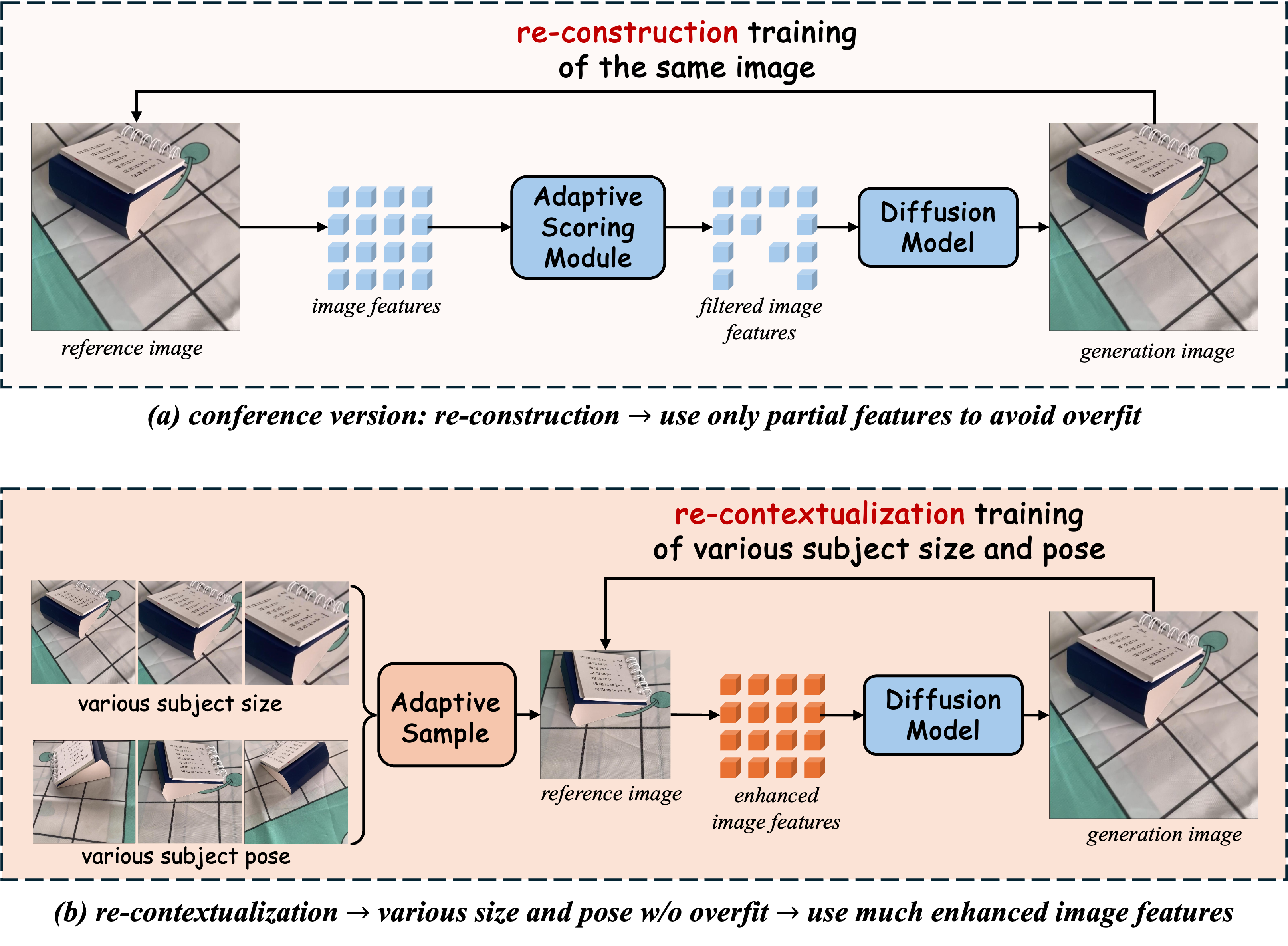}
  \caption{Schematic comparison. We \emph{completely redesign the paradigm}: the conference version~\cite{huang2024realcustom} adopts \emph{reconstruction training}, which restricts fine-grained image features to avoid overfitting. Our new \emph{re-contextualization training} introduces references with diverse subject sizes and poses, effectively preventing overfitting and enabling the use of richer image features, leading to simultaneous improvements in both similarity and controllability.}
  \label{method_discuss_conference}
\end{figure}


\textbf{{(1) New training recipe and subject representation enabled by newly designed re-contextualization training:}} (i) The \emph{Curriculum Training Recipe (CTR)}, which curates an ``easy-to-hard" data for subject images, allowing for a gradual increase in controllability. (ii) The \emph{Cross-layer Cross-scale Projector (CCP)}, which adaptively fuses multi-layer and multi-scale image features for enhanced subject similarity.

\textbf{(2) New guidance mask algorithm for more accurate disentanglement:} (i) \emph{More Spatial Accurate:} We introduce a self-attention augmented cross-attention calculation method, which reduces the noise of naive cross-attention by incorporating per-pixel correlation. (ii) \emph{More Temporal Stable:} We propose an ``early stop" regularization to stabilize the guidance mask in later diffusion steps, which also accelerates the generation.


\textbf{(3) More comprehensive experiments:} (i) \emph{Advanced backbones}: Beyond the conference version’s evaluation on SD-v1.5, we further validate our paradigm on the advanced SDXL backbone. (ii) \emph{Advanced tasks}: We introduce a novel multiple adaptive mask guidance algorithm, extending RealCustom++ to support multiple subject customization across diverse tasks.

\textbf{(4) Higher performance:} Compared to our conference version, we achieve improvements of 6.06\% in text controllability and 2.32\% in subject similarity simultaneously, and a significant 56.6\% improvement on generation quality. 

\subsection{Text-to-Image Customization}


Existing text-to-image customization methods, following the pseudo-word paradigm, can be classified as either optimization-based~\cite{gal2022image, ruiz2023dreambooth, kumari2023multi, alaluf2023neural, voynov2023p+, daras2022multiresolution, liu2023cones, dong2022dreamartist} or optimization-free~\cite{wei2023elite, li2024blip, shi2023instantbooth, gal2023designing, jia2023taming, chen2023dreamidentity}.


\textbf{\emph{Optimization-based customization stream}.} Textual Inversion~\cite{gal2022image} first represented a given subject with new ``words'' in the embedding space of a frozen text-to-image model. DreamBooth~\cite{ruiz2023dreambooth} used a rare token as the pseudo word and fine-tuned the entire diffusion model to enhance similarity. Custom Diffusion~\cite{kumari2023multi} identified and optimized only key parameters (\emph{e.g.}, key and value projection layers) for customization. P+~\cite{voynov2023p+} extended textual inversion by learning per-layer pseudo words for faster convergence. Cones~\cite{liu2023cones} optimized residual token embeddings per subject. Building on these approaches, subsequent works~\cite{hao2023vico, hua2023dreamtuner} incorporated finer image patch features via additional image attention modules, improving subject similarity but increasing overfitting risk. The primary limitation of this line of work is the lengthy optimization time (minutes to hours) and the need to store fine-tuned weights for each subject, resulting in significant computational and memory overhead that limits practical use.

\textbf{\emph{Optimization-free customization stream}}. To deal with the cumbersome requirement of per-subject optimization, optimization-free customization methods have emerged, typically training an image encoder to project subjects into pseudo words using object-level datasets (\emph{e.g.}, OpenImages~\cite{kuznetsova2020open}, FFHQ~\cite{karras2019style}). ELITE~\cite{wei2023elite} introduced a learning-based encoder with global and local mapping networks for rapid subject customization. BLIP-Diffusion~\cite{li2024blip} employs a multimodal encoder for improved subject representation. InstantBooth~\cite{shi2023instantbooth} integrates adapter layers into pre-trained diffusion models, but is limited to a few categories (\emph{i.e.}, human or cat). Subject Diffusion~\cite{ma2023subject} leverages a large-scale dataset with detection boxes, segmentation masks, and text descriptions, but generated subjects often lack pose diversity due to a simple reconstruction objective. Some works~\cite{li2023photomaker, ye2023ip, wang2024instantid, chen2023dreamidentity} focus on human ID-specific customization, such as PhotoMaker~\cite{li2023photomaker} stacking ID embeddings as pseudo words and InstantID~\cite{wang2024instantid} introducing IdentityNet with weak spatial constraints. However, these methods still suffer from limited pose diversity, with generated faces rigidly facing the camera.

Recently, \emph{Pan et al.}~\cite{pan2024locate} proposed LAR-Gen, which inpaints specific subjects into user-defined masked regions. We differ from LAR-Gen in: (1) \emph{Task}: LAR-Gen requires user masks for inpainting, while our method enables free-form customized image generation without mask input for greater convenience.  (2) \emph{Objective}: LAR-Gen seeks seamless integration of inpainted subjects with the source image, while our goal is to disentangle text and image conditions to jointly optimize controllability and similarity. (3) \emph{Method}: Our framework and all modules are different from LAR-Gen. LAR-Gen employs an auxiliary diffusion U-Net as the image encoder for masked regions and concatenates the reference image with noise as inputs. We introduce a Cross-layer Cross-scale Projector for fine-grained subject features, a Curriculum Training Recipe to systematically control the size and pose variations of input images during training, and a dual-branch inference framework to disentangle the inference scopes of text and image conditions.



Meanwhile, ACE~\cite{han2024ace} proposes an impressive and promising unified framework for multiple image generation and editing tasks. However, it does not yet support text-to-image customization. While ACE focuses on developing a unified framework and addressing key challenges such as unifying various conditioning formats, our work specifically targets text-to-image customization tasks and aims to push the upper bound of text-to-image customization performance. We believe that advancements in text-to-image customization can provide high-quality, task-specific data to further enhance the development of more powerful unified image generation frameworks.


\subsection{Multiple subject customization}
Although most text-to-image customization methods focus on single-subject scenarios, interest in multiple-subject customization is increasing. This task can be categorized as: (1) decoupling multiple subjects within a single reference image (\emph{One2Many}), and (2) learning each subject from its own reference image and composing them into one output (\emph{Many2Many}). Existing approaches typically extend the pseudo-word paradigm from single-subject customization by assigning distinct pseudo words to each subject and developing algorithms to disentangle their representations.

\textbf{\emph{One2Many: decoupling multiple subjects from a single reference image.}} Break-a-scene~\cite{avrahami2023break} first proposed extracting a distinct pseudo word for each subject by augmenting the input image with user-provided or segmentation-generated masks indicating target subjects. Subsequent works~\cite{jin2023image, hao2023vico} automated mask generation using the cross-attention maps of learned pseudo words, applying either fixed or Otsu~\cite{otsu1975threshold} thresholding. DisenDiff~\cite{zhang2024attention} introduced attention calibration to distribute attention across concepts and achieve disentanglement without masks. However, these methods often struggle to handle the reference background in the generated results.

\textbf{\emph{Many2Many: composing multiple subjects from multiple reference images.}} This line of customization methods primarily addresses inter-confusion between different learned subject pseudo words. Custom Diffusion~\cite{kumari2023multi} proposed joint training on multiple concepts with constrained optimization to merge them, but typically requires over ten reference images per subject for convergence. Mix-of-Show~\cite{gu2024mix} employs an embedding-decomposed LoRA~\cite{hu2021lora} for subject-specific optimization. A similar idea is adopted in MultiBooth\cite{zhu2024multibooth} and MS-Diffusion\cite{wang2024ms}, where each subject is bounded by pre-defined bounding boxes to define its specific generation area.



\subsection{Curriculum Learning}

Curriculum learning~\cite{bengio2009curriculum, wang2021survey, wang2024efficienttrain++, wu2024time} is a training strategy that organizes data from ``easy'' to ``hard'' to mimic human learning, aiming to improve model performance and accelerate convergence. It has been widely applied in areas such as entity relation extraction~\cite{xu2021entity}, with most methods focusing on data difficulty estimation~\cite{wei2016stc, tudor2016hard} or training schedulers~\cite{cirik2016visualizing, matiisen2019teacher}. However, its application in image generation remains underexplored. In this work, we address this gap by organizing subject image data to gradually shift the customization task from simple reconstruction to challenging re-contextualization, bridging the gap between training and inference and enabling open-domain coherent subject generation.
\section{Methodology: RealCustom++}

In this study, we focus on the most general customization scenario, \emph{i.e.}, generating new high-quality images for a given subject based on a single reference image and following the given text. The generated subject may vary in location, pose, style, \emph{etc.}, yet it should maintain high-quality similarity with the reference. The remaining parts of the generated images should consistently adhere to the given text, ensuring high-quality controllability.

In \cref{method_analysis_and_empirical_validation}, we analyze and empirically validate the dual-optimum paradox in existing pseudo-word paradigms, motivating our proposed real-word paradigm, RealCustom++. Preliminaries are introduced in \cref{preliminaries}. The training and inference frameworks of RealCustom++ are detailed in \cref{paradigm_train} and \cref{paradigm_inference}. Finally, we describe the extension to multiple-subject customization in \cref{paradigm_inference_multiple}.

\subsection{Analysis \& Empirical Validation}
\label{method_analysis_and_empirical_validation}



\begin{figure}
  \centering
  \includegraphics[width=1.0\linewidth]{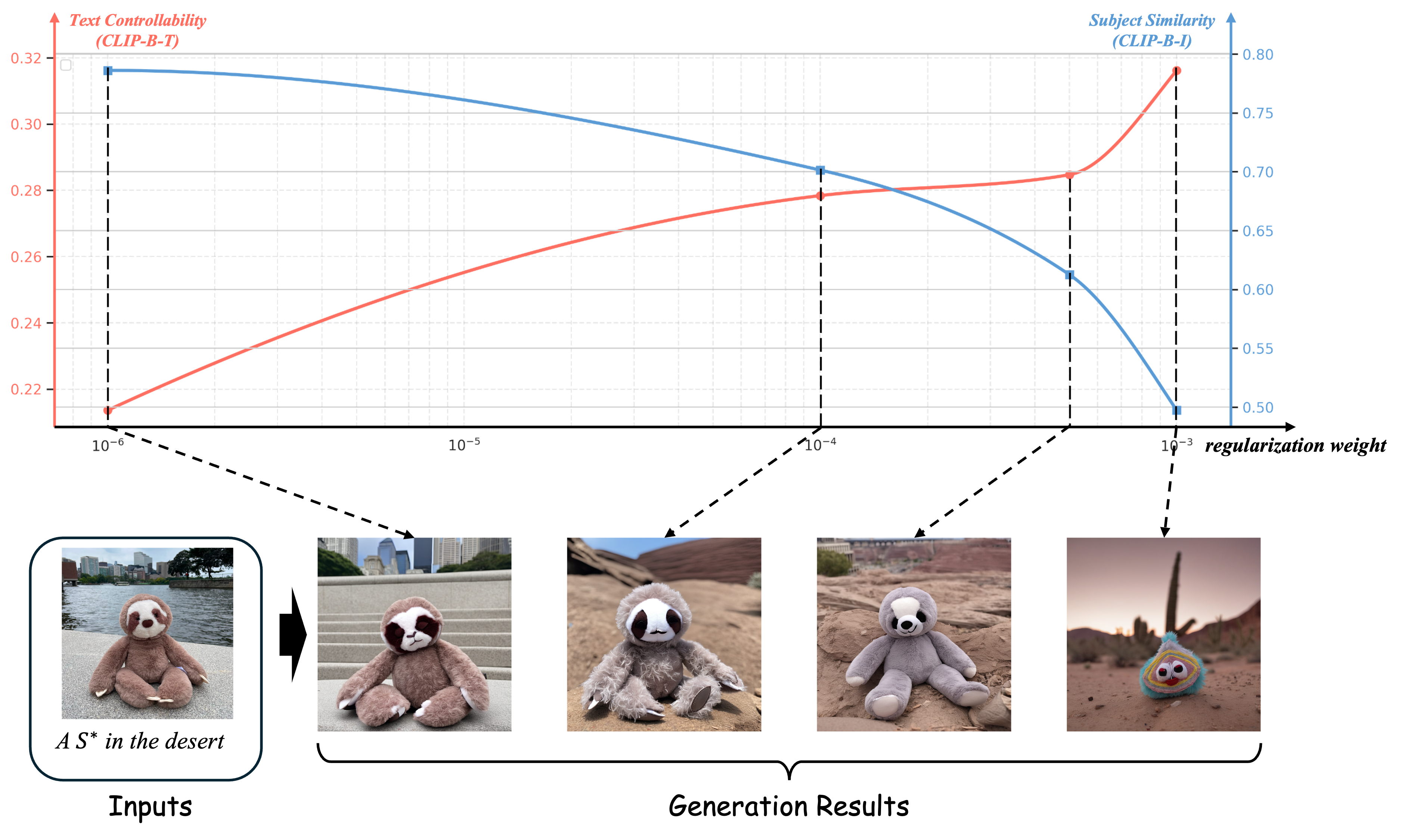}
  \caption{Demonstration of the trade-off between subject similarity and text controllability in existing pseudo-word paradigms (illustrated with a representative pseudo-word approach, \emph{i.e.}, Textual Inversion~\cite{gal2022image}): increasing regularization weight reduces subject similarity but improves text controllability, revealing a dual-optimum paradox where both cannot be maximized simultaneously.}
  \label{method_illustration_pseudo_word}
\end{figure}

\begin{table}[!t]
    \caption{Illustration of semantic conflict}
    \label{method_semantic_similarity}
    \centering
    \begin{tabular}{cc}
    \hline
    Regularization Weight of $S^*$ & Semantic Similarity of ``desert" $\uparrow$ \\
    \hline
    0 & 0.2175 \\
    1e-4 & 0.2712 \\
    5e-4 & 0.2772 \\
    1e-3 & 0.5879 \\
    \hline
    \end{tabular}
    \begin{tablenotes}
    \footnotesize
    \item We train pseudo words $S^*$ under varying regularization losses to examine how they cause real words (\emph{e.g.}, ``desert'') to deviate from their original semantics. Semantic similarity is measured using cosine distance. We observe that a lower regularization weight for the pseudo word leads to reduced semantic similarity for the real word ``desert'', indicating more pronounced semantic drift and diminished text controllability. This aligns with the decrease in textual controllability shown in \cref{method_illustration_pseudo_word}.
    \end{tablenotes}
\end{table}

\textbf{Observations.} Existing customization methods typically adopt the pseudo-word paradigm, \emph{i.e.}, they first learn pseudo words to represent the visual subject and then combine these with other texts as a unified condition for image generation. The embedding of these pseudo words, denoted as $v_{*}$, are optimized using both a diffusion loss and a regularization loss:

\begin{equation}
    L = L_{\text{diffusion}} + \lambda L_{\text{regularization}},
\end{equation}
Here, the diffusion loss $L_{\text{diffusion}}$ encourages $v_{*}$ to capture subject characteristics, while the regularization loss $L_{\text{regularization}}$ can be either an $\ell_1$ regularization to constrain the $\ell_1$-norm of $v_{*}$~\cite{wei2023elite, jin2023image, li2023photomaker, hao2023vico}, or a prior preservation regularization using regularization images~\cite{ruiz2023dreambooth, kumari2023multi}. The parameter $\lambda$ is a manually set hyperparameter balancing these losses. To illustrate the trade-off between subject similarity and text controllability, we train a representative pseudo-word method (\emph{i.e.}, Textual Inversion~\cite{gal2022image}) with varying $\lambda$, as shown in \cref{method_illustration_pseudo_word}. We observe that increasing $\lambda$ decreases subject similarity but improves text controllability, revealing a dual-optimum paradox, \emph{i.e.}, both objectives cannot be maximized simultaneously. As a result, most methods adopt a moderate $\lambda$ (\emph{e.g.}, $10^{-4}$) to achieve a practical balance.

\textbf{Diagnosis \& Analysis.} In this study, we argue that the dual-optimum paradox stems from inherent \textbf{\emph{conflicts}} and \textbf{\emph{entanglements}} between pseudo words and the given text. First, \textbf{\emph{conflicts}} arise from the non-homologous nature of pseudo and real word representations. Pseudo words are learned by reconstructing subject images in a visual-only manner (\emph{i.e.}, via diffusion loss), without contextual alignment to other text words. In contrast, real words are learned through large-scale linguistic or multimodal pre-training (\emph{e.g.}, T5~\cite{raffel2020exploring}, CLIP~\cite{radford2021learning}), resulting in semantically coherent contexts. As a result, pseudo words often conflict with real text words, as their representations are context-independent. Increasing the influence of pseudo words leads real words to deviate from their original semantics, reducing controllability, while decreasing their influence reduces subject similarity. In \cref{method_semantic_similarity}, we design a diagnostic experiment to illustrate how pseudo-words can cause real words to deviate from their original semantics. The experiment proceeds as follows:

\noindent 1) The prompt ``A toy in the desert'' is encoded using the text encoder, with the embedding of ``desert'' as the ground truth.

\noindent 2) For each experiment with different regularization weights, we encode the prompt ``A $S^*$ in the desert'' with the same text encoder and extract the embedding of ``desert''.

\noindent  3) We compute the cosine distance between the ``desert'' embedding obtained in step (2) and the ground truth from step (1) to quantify semantic similarity and assess the extent of semantic drift in the real word ``desert'' caused by the introduction of the pseudo-word $S^*$.

As shown in \cref{method_semantic_similarity}, a smaller regularization weight $\lambda$ for the pseudo-word $S^*$ results in lower semantic similarity for the real word ``desert'', indicating more severe semantic drift and diminished text controllability. This observation is consistent with the generation results in \cref{method_illustration_pseudo_word}, where a smaller regularization weight $\lambda$ also shows reduced text controllability.

On the other hand, (2) \textbf{\emph{entanglements}} arise from the overlapping influence of the given text and subjects. In the pseudo-word paradigm, both pseudo and real words jointly control generation via cross-attention, updating each image region as a weighted sum of all tokens. This causes subjects to influence all regions indiscriminately, regardless of relevance. As a result, increasing the impact of pseudo-words to enhance subject similarity in relevant regions also amplifies their influence in irrelevant regions, causing these regions to follow the subject rather than the text and thus reducing controllability, and vice versa. As shown in \cref{method_illustration_pseudo_word}, text controllability is primarily determined by the alignment of non-subject regions with the textual description (\emph{e.g.}, ``desert''), while subject similarity depends on the alignment of the subject region with the reference image. Specifically, increasing the regularization weight improves text controllability by enhancing alignment in non-subject regions, but simultaneously reduces subject similarity by weakening alignment in the subject region.


\textbf{Motivations.} The above analysis motivates RealCustom++, which addresses the dual-optimum paradox with two key insights. (1) To eliminate conflicts, we represent the subject using a real text word and generate a guidance image with a non-conflicting layout, identifying subject-relevant regions at each generation step. (2) To disentangle optimization objectives, we constrain the influence of visual conditions to subject-relevant regions, while subject-irrelevant regions are purely controlled by the text. \emph{This allows for the simultaneous optimization of subject similarity in subject-relevant regions and text controllability in subject-irrelevant regions.}

\begin{figure*}
  \centering
  \includegraphics[width=0.9\linewidth]{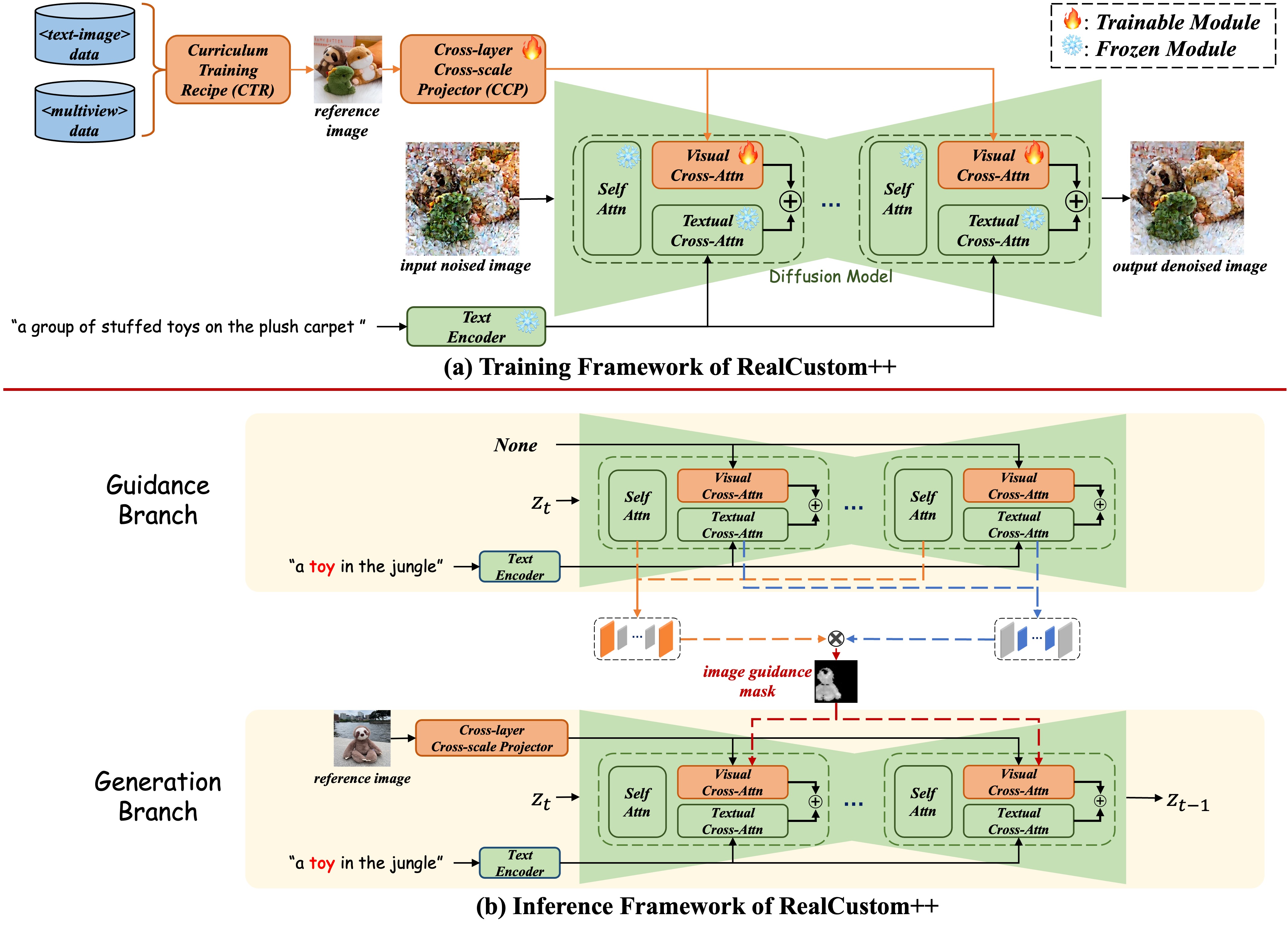}
  \caption{(a) Illustration of the RealCustom++ training framework, which learns general alignment between vision conditions and all text words. This is enabled by the \emph{Cross-layer Cross-Scale Projector (CCP)} for robust, fine-grained subject representation, and a \emph{Curriculum Training Recipe (CTR)} to adapt subjects to diverse poses and sizes. Subject representations are injected into the diffusion model by extending textual cross-attention with an additional visual cross-attention in each block. (b) Illustration of the RealCustom++ inference framework at generation step $t$, where \emph{Adaptive Mask Guidance (AMG)} customizes the generation for a specific real word (\emph{e.g.}, ``toy''). This involves two branches per generation step: a Guidance Branch that constructs the image guidance mask, and a Generation Branch that uses this mask to preserve uncontaminated subject-irrelevant regions.}
  \label{framework}
\end{figure*}

\subsection{Preliminaries}
\label{preliminaries}

Our paradigm is built upon Stable Diffusion~\cite{rombach2022high, podell2023sdxl}, which comprises an autoencoder and a conditional UNet~\cite{ronneberger2015u} denoiser. Given an image $\boldsymbol{x} \in \mathbb{R}^{H \times W \times 3}$, the autoencoder encoder $\mathcal{E}(\cdot)$ maps it to a latent space $\boldsymbol{z} = \mathcal{E}(\boldsymbol{x}) \in \mathbb{R}^{h \times w \times c}$, where $f = H/h = W/w$ is the downsampling factor and $c$ the latent channel dimension. The decoder $\mathcal{D}(\cdot)$ reconstructs the image as $\mathcal{D}(\mathcal{E}(\boldsymbol{x})) \approx \boldsymbol{x}$. The conditional denoiser $\epsilon_\theta(\cdot)$ operates in latent space and is conditioned on text features $\boldsymbol{f_{ct}} = \tau_{\text{text}}(y)$, where $\tau_{\text{text}}(\cdot)$ is the pre-trained CLIP text encoder~\cite{radford2021learning}. The denoiser is trained with:
\begin{equation}
    L:=\mathbb{E}_{\boldsymbol{z} \sim \mathcal{E}(\boldsymbol{x}), \boldsymbol{f_y}, \boldsymbol{\epsilon} \sim \mathcal{N}(\boldsymbol{0},\boldsymbol{\text{I}}), t}\left[\left\|\boldsymbol{\epsilon} -\epsilon_\theta\left(\boldsymbol{z_t}, t, \boldsymbol{f_{ct}}\right)\right\|_2^2\right],
\end{equation}
where $\boldsymbol{\epsilon}$ denotes for the unscaled noise and $t$ is the timestep. $\boldsymbol{z_t}$ is the latent vector that noised according to $t$.
During inference, random Gaussian noise $\boldsymbol{z_T}$ is iteratively denoised to $\boldsymbol{z_0}$, and then reconstructed as $\boldsymbol{x}' = \mathcal{D}(\boldsymbol{z_0})$. Text conditioning in Stable Diffusion is implemented via textual cross-attention:
\begin{equation}
    \label{attention}
    \text{Attention}(\boldsymbol{Q},\boldsymbol{K},\boldsymbol{V}) = \text{Softmax}(\boldsymbol{Q}\boldsymbol{K}^{\top})\boldsymbol{V},
\end{equation}
where $\boldsymbol{Q} = \boldsymbol{W_Q} \cdot \boldsymbol{f_i}$, $\boldsymbol{K} = \boldsymbol{W_K} \cdot \boldsymbol{f_{ct}}$, and $\boldsymbol{V} = \boldsymbol{W_V} \cdot \boldsymbol{f_{ct}}$, with $\boldsymbol{W_Q}, \boldsymbol{W_K}, \boldsymbol{W_V}$ as projection weights. $\boldsymbol{f_i}$ and $\boldsymbol{f_{ct}}$ denote latent image and text features, respectively. The latent image features are updated using the attention output.

\subsection{Training Framework}
\label{paradigm_train}

Given a reference image $\boldsymbol{x} \in \mathbb{R}^{3 \times H \times W}$, where $H$ and $W$ are the image height and width, respectively, we employ the proposed Cross-layer Cross-scale Projector (CCP) to extract a visual condition $\boldsymbol{f_{ci}} \in \mathbb{R}^{n_{\text{image}} \times c_{\text{image}}}$, as shown in \cref{framework}(a). Here, $n_{\text{image}}$ and $c_{\text{image}}$ denote the number of tokens and feature dimension. We then extend the textual cross-attention in pre-trained diffusion models by introducing an parallel visual cross-attention. Specifically, \cref{attention} is reformulated as:
\begin{multline}
    \label{dual_attention}
    \text{Attention}(\boldsymbol{Q},\boldsymbol{K},\boldsymbol{V},\boldsymbol{K_i},\boldsymbol{V_i}) = \\
    \text{Softmax}(\boldsymbol{Q}\boldsymbol{K}^{\top})\boldsymbol{V} + \text{Softmax}(\boldsymbol{Q}\boldsymbol{K_i}^{\top})\boldsymbol{V_i},
\end{multline}
where the new key $\boldsymbol{K_i} = \boldsymbol{W_{Ki}} \cdot \boldsymbol{f_{ci}}$, value $\boldsymbol{V_i} = \boldsymbol{W_{Vi}} \cdot \boldsymbol{f_{ci}}$ are added. $\boldsymbol{W_{Ki}}$ and $\boldsymbol{W_{Vi}}$ are weight parameters. During training, only the CCP module and projection layers $\boldsymbol{W_{Ki}}, \boldsymbol{W_{Vi}}$ in each attention block are trainable. To better align the vision condition with the textual semantics and avoid the training degrading to a naive copy-paste reconstruction, we further propose a curriculum training recipe with a novel ``easier-to-harder" subject image data procedure, allowing the subject to be generated in various poses and sizes with coherently.

\subsubsection{Cross-layer Cross-scale Projector}

Existing works~\cite{huang2024realcustom, wei2023elite, shi2023instantbooth, ye2023ip, wang2024instantid} typically encode subject images using the last hidden states of pretrained ViT encoders (\emph{e.g.}, CLIP~\cite{radford2021learning}), which are trained on discriminative tasks such as contrastive learning~\cite{radford2021learning}. These encoders produce semantically rich representations that facilitate visual-textual alignment, but they exhibit two main limitations:


\noindent \textbf{\emph{(i) Insufficient structural robustness}} Due to the highly semantic nature of deep features, they exhibit limited capacity to preserve low-level shape and structural information, especially for complex subjects (\emph{e.g.}, buildings).

\noindent \textbf{\emph{(ii) Limited detail richness}}. Most existing image encoders are typically trained on low resolutions (\emph{e.g.}, 224 or 384). Consequently, the given subjects must be resized to these low resolutions, resulting in the loss of numerous subject details.

To address these limitations in existing subject representations, we propose a novel cross-layer cross-scale projector. The key idea is to enhance the subject's low-resolution deep features with its shallow and high-resolution counterparts, without compromising their initial alignment with the text condition or the efficiency of short token length.

\begin{figure}
  \centering
  \includegraphics[width=0.9\linewidth]{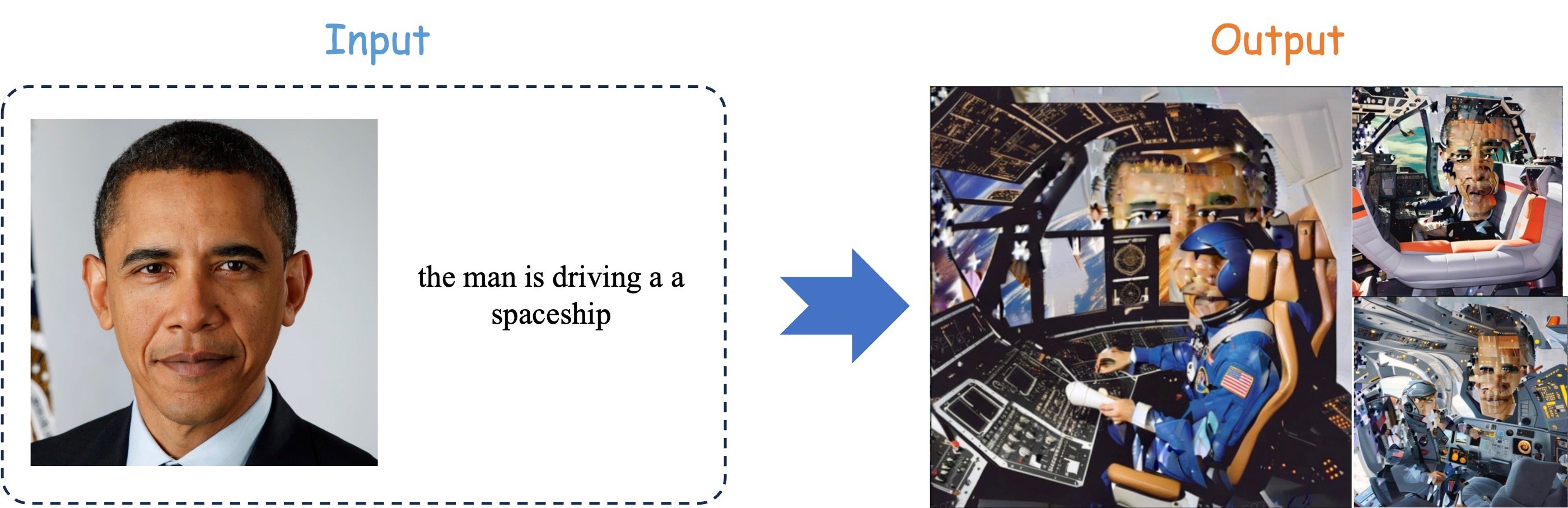}
  \caption{Illustration of a typical failure when naively concatenating shallow features with deep ones, resulting in a ``copy-paste" problem that disrupts the alignment between text and deep image features.}
  \label{failure_case}
\end{figure}

\textbf{\emph{Cross-layer attention to enhance structural robustness:}} Unlike deep features from pretrained image encoders, shallow features retain more low-level structural information~\cite{hua2023dreamtuner}. Simply concatenating shallow and deep features disrupts the alignment of deep features, leading to ``copy-paste'' artifacts (\cref{failure_case}). We hypothesize that shallow features, being overly low-level, dominate training and cause direct transfer of reference image parts. To address this, we design cross-layer attention, allowing deep features to query shallow ones for structural robustness while maintaining their dominance.

\begin{figure*}
  \centering
  \includegraphics[width=0.9\linewidth]{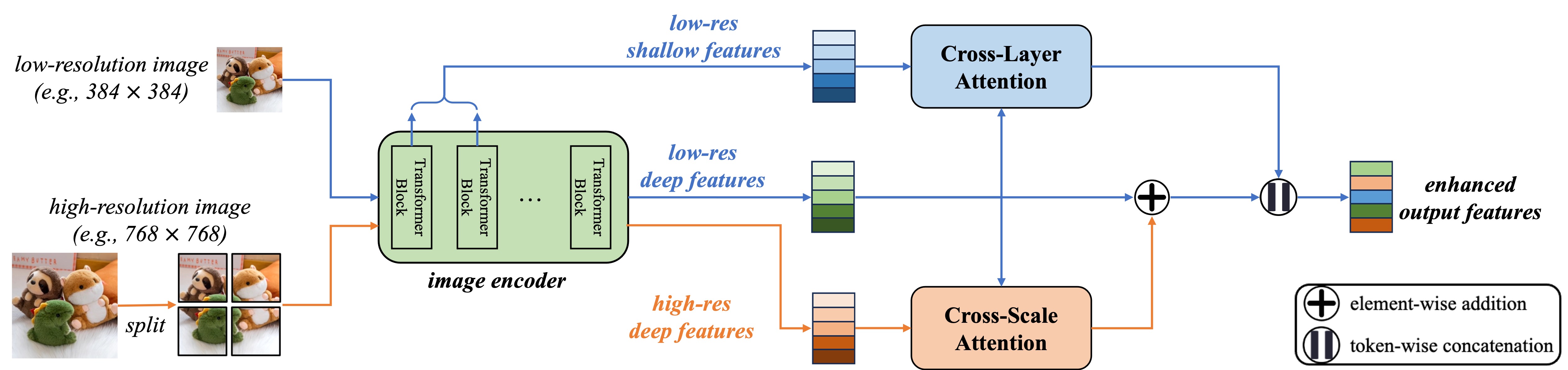}
  \caption{Illustration of the proposed Cross-Layer Cross-Scale Projector.}
  \label{projector}
\end{figure*}

Mathematically, let the low-resolution deep image features from the last hidden state be $\boldsymbol{f_{\text{deep}}} \in \mathbb{R}^{n_{\text{image}} \times c_{0}}$. We select $L$ layers of shallow image features, denoted as $\boldsymbol{f_{\text{shallow}}^l} \in \mathbb{R}^{n_{\text{image}} \times c_{0}}$, where $l \in [0, L-1]$ and $c_{0}$ is the feature dimension of the pretrained image encoder. These shallow features are concatenated along the token dimension to form $\boldsymbol{f_{\text{shallow}}} \in \mathbb{R}^{(n_{\text{image}} \times L) \times c_{0}}$. The cross-layer attention is then defined as:

\begin{equation}
    \boldsymbol{f_{\text{shallow}}^{'}} =
        \text{Softmax}(\boldsymbol{Q^{s}}\boldsymbol{K^{s}}^{\top})\boldsymbol{V^{s}}
    \in \mathbb{R}^{n_{\text{image}} \times c_{0}},
\end{equation}
where $\boldsymbol{Q^{s}} = \boldsymbol{W_{Qs}} \cdot \boldsymbol{f_{\text{deep}}}, \boldsymbol{K^{s}} = \boldsymbol{W_{Ks}} \cdot \boldsymbol{f_{\text{shallow}}}, \boldsymbol{V^{s}} = \boldsymbol{W_{Vs}} \cdot \boldsymbol{f_{\text{shallow}}}$. Here, $\boldsymbol{W_{Qs}}, \boldsymbol{W_{Ks}}, \boldsymbol{W_{Vs}} \in \mathbb{R}^{c_0 \times c_0}$ are learnable projection matrix.
The output features are further projected into the desired dimension as:
\begin{align}
    \boldsymbol{f_{\text{shallow}}^{'}} = \text{MLP}(\boldsymbol{f_{\text{shallow}}^{'}}) \in \mathbb{R}^{n_{\text{image}} \times c_{\text{image}}}
\end{align}

\textbf{\emph{Cross-scale attention to enrich subject details:}}
As shown in \cref{projector}, we introduce a high-resolution image stream to enhance subject detail. The subject image is first resized to twice the encoder's target resolution (\emph{e.g.}, $768^2$ for a ViT encoder~\cite{zhai2023sigmoid} pretrained on $384^2$). The $768^2$ image is divided into four $384^2$ patches, which are processed by the pretrained encoder. The resulting local features are concatenated along the token dimension to form $\boldsymbol{f_{\text{high}}} \in \mathbb{R}^{(4 n_{\text{image}}) \times c_0}$. Directly using these quadruple-length features significantly increases the training cost (\emph{i.e.}, experimentally, the maximum batch size is limited to 1 when directly feeding these features into the visual cross-attention). To address this, we implement a cross-scale attention mechanism to integrate high-resolution local features with the original low-resolution global features without increasing the token length:

\begin{equation}
    \boldsymbol{f_{\text{high}}^{'}} = \text{Softmax}(\boldsymbol{Q^h}\boldsymbol{K^h}^{\top})\boldsymbol{V^h} \in \mathbb{R}^{n_{\text{image}} \times c_{0}},
\end{equation}
where $\boldsymbol{Q^h} = \boldsymbol{W_{Qh}} \cdot \boldsymbol{f_{\text{deep}}}, \boldsymbol{K^h} = \boldsymbol{W_{Kh}} \cdot \boldsymbol{f_{\text{high}}}, \boldsymbol{V^h} = \boldsymbol{W_{hv}} \cdot \boldsymbol{f_{\text{high}}}$. 
Here, $\boldsymbol{W_{Qh}}, \boldsymbol{W_{Kh}}, \boldsymbol{W_{hv}} \in \mathbb{R}^{c_0 \times c_0}$ are learnable projection matrices.
Similarly,

\begin{align}
    \boldsymbol{f_{\text{high}}^{'}} = \text{MLP}(\boldsymbol{f_{\text{high}}^{'}}) \in \mathbb{R}^{n_{\text{image}} \times c_{\text{image}}}
\end{align}



\textbf{\emph{Feature combination:}} After obtaining the structure-enhanced features $\boldsymbol{f_{\text{shallow}}^{'}}$ and detail-enhanced features $\boldsymbol{f_{\text{high}}^{'}}$, we combine them with the original low-resolution deep features $\boldsymbol{f_{\text{deep}}}$ via token-wise concatenation and element-wise addition, respectively:

\begin{equation}
    f_{ci} = \left[ f_{\text{shallow}}^{\prime} ;\, \text{MLP}(f_{\text{deep}}) \oplus f_{\text{high}}^{\prime} \right] \in \mathbb{R}^{n_{\text{image}} \times c_{\text{image}}}.
\end{equation}
Here, MLP denotes a multi-layer perceptron that projects $\boldsymbol{f_{\text{deep}}}$ to the target output dimension $c_{\text{image}}$. $[\cdot\,;\,\cdot]$ and $\oplus$ indicate token-wise concatenation and element-wise addition, respectively. The design rationale is that $\boldsymbol{f_{\text{high}}^{'}}$ and $\boldsymbol{f_{\text{deep}}}$ represent features at the same semantic level (\emph{i.e.}, last hidden states), so element-wise addition preserves consistency. In contrast, $\boldsymbol{f_{\text{shallow}}^{'}}$ encodes features from different levels, and token-wise concatenation enables the model to adaptively select information across levels in \cref{dual_attention}.

\subsubsection{Curriculum Training Recipe}
\label{Curriculum_Training_Recipe}
\begin{figure}
  \centering
  \includegraphics[width=0.9\linewidth]{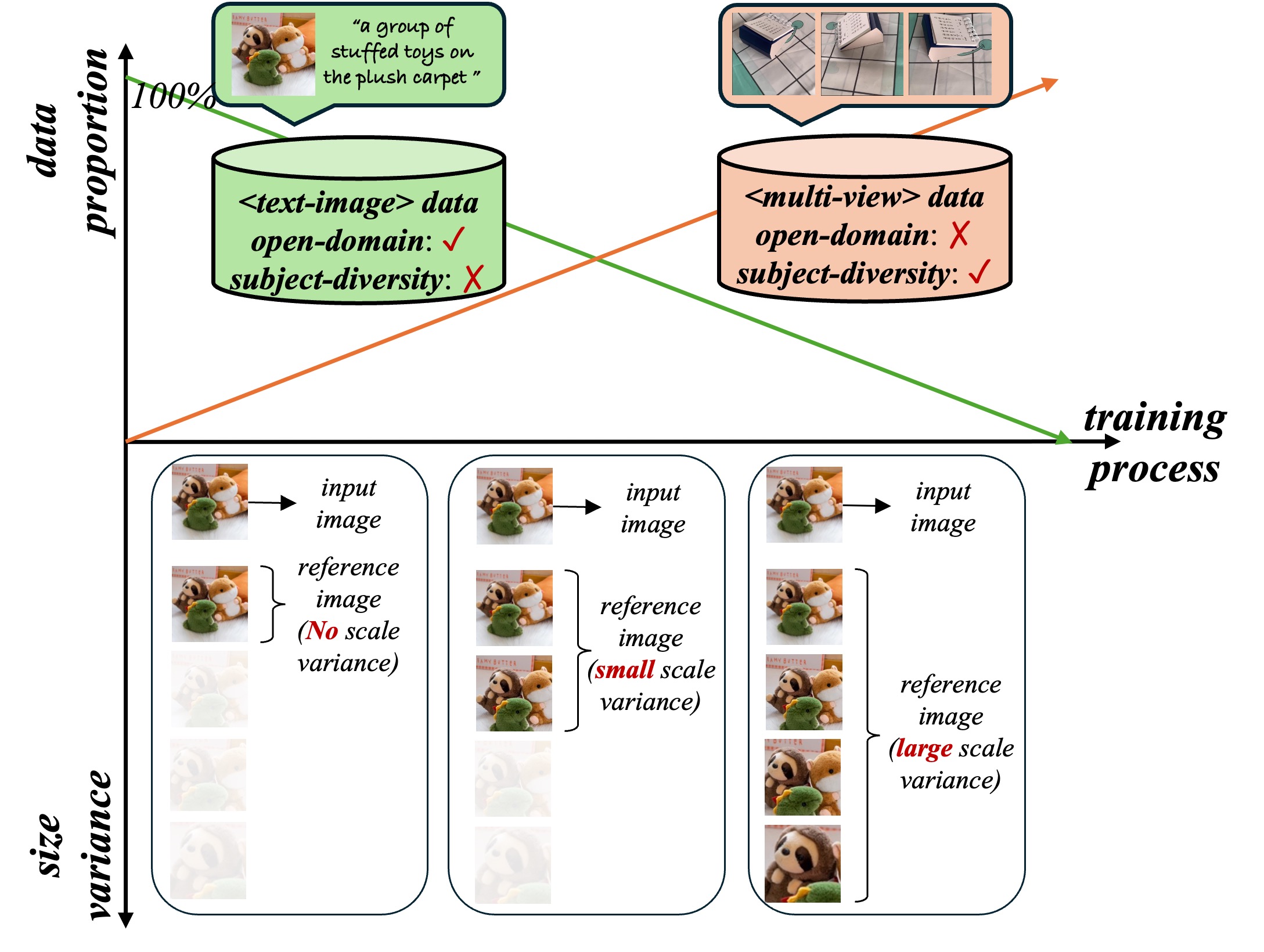}
  \caption{
  Illustration of the proposed curriculum training recipe (CTR): To enhance pose diversity while preserving open-domain generation, RealCustom++ is trained on a mixture of generic \texttt{<text-image>} data and \texttt{<multiview>} data~\cite{yu2023mvimgnet}. The dataset proportions are progressively adjusted, starting with a higher ratio of open-domain data and gradually increasing the share of multiview data to improve pose generalization. Simultaneously, the model is exposed to reference images with increasing size variance, enabling rapid initial convergence and subsequent generalization to diverse subject sizes.}
  \label{framework_train_recipe}
\end{figure}

The key to open-domain subject customization is training on open-vocabulary datasets to ensure generalization to unseen subjects. In our conference version, we trained the model on a generic \texttt{<text-image>} dataset (\emph{i.e.}, Laion-5B~\cite{schuhmann2022laion}), using the same images as both visual conditions and inputs to the diffusion denoiser. This led to a train-inference gap: during training, generated subjects matched the pose and size of the reference images, whereas during inference, subjects needed to adapt to diverse poses and sizes specified by text prompts. 
We emphasize that appropriate training data settings are essential for robust customization. As shown in \cref{framework_train_recipe}, we propose a novel curriculum training strategy that enables the model to learn subject customization across a range of poses and sizes without additional architectural modifications.


\textbf{\emph{Adaption to diverse subject pose:}} 
We expand the training data used in our conference version~\cite{huang2024realcustom} to include a mixture of a generic \texttt{<text-image>} dataset (\emph{e.g.}, LAION~\cite{schuhmann2022laion}) and a \texttt{<multiview>} dataset (\emph{e.g.}, MVImageNet~\cite{yu2023mvimgnet}). The former covers a broad spectrum of open-domain subjects but lacks diversity in subject poses and views, while the latter offers multiple poses and views per subject but is restricted to a limited set of categories (\emph{i.e.}, 238 classes mainly consisting of daily necessities). We propose to progressively increase the proportion of \texttt{<multiview>} data while decreasing the proportion of \texttt{<text-image>} data during training. This strategy initially exposes the model to extensive open-domain data, enabling rapid convergence and generalization through a naive reconstruction task. As the proportion of \texttt{<multiview>} data increases, the training task gradually shifts toward diverse view synthesis, thereby improving the model's ability to handle varied subject poses without sacrificing generalization.

Specifically, given the total training steps $S_{\text{total}}$ and the current training step $S_{\text{cur}}$, the probability of using the \texttt{<text-image>} dataset and \texttt{<multiview>} data for the current training step is $P^{\text{generic}}_{\text{cur}}$ and $P^{\text{multiview}}_{\text{cur}}$, respectively:

\begin{align}
    P^{\text{multiview}}_{\text{cur}} = \frac{S_{\text{cur}}}{S_{\text{total}}},
    P^{\text{generic}}_{\text{cur}} = 1 - P^{\text{multiview}}_{\text{cur}}.
\end{align}

\begin{figure}
  \centering
  \includegraphics[width=0.9\linewidth]{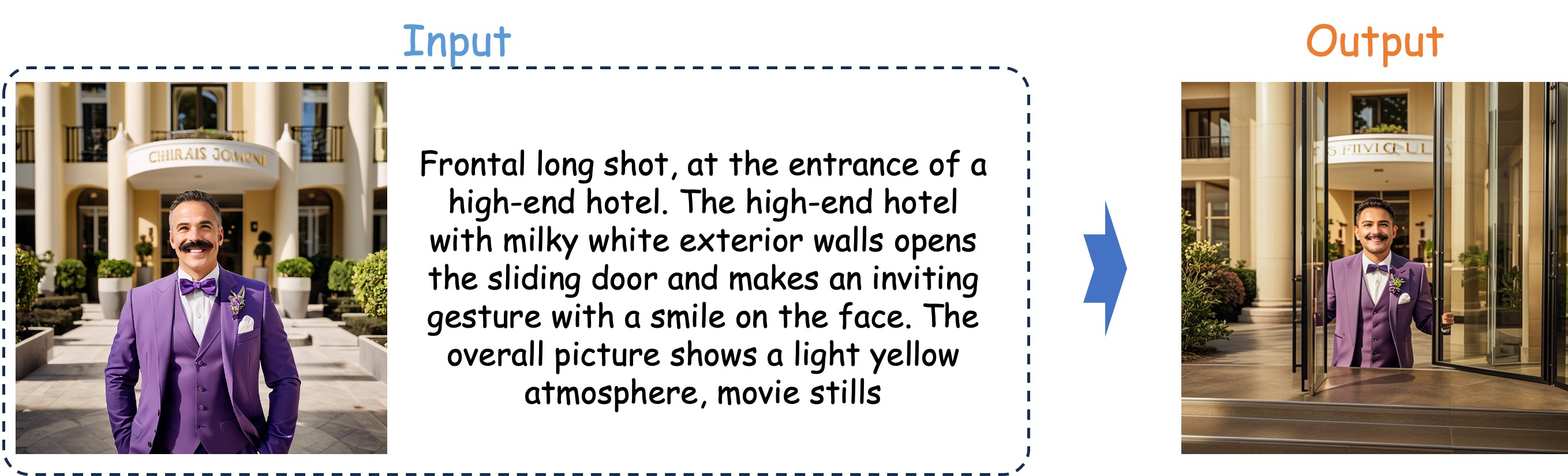}
  \caption{Illustration of typical failure cases arising from training without subject size adaptation, which leads to incoherence between subject regions and text-controlled, subject-irrelevant regions. This often manifests as disproportionate figures, such as a ``large head with a small body'' or ``half body''.}
  \label{failure_case2}
\end{figure}

\textbf{\emph{Adaption to diverse subject size:}} 
While multiview data improves adaptability to diverse poses and views, we still observe incoherence between subject regions and text-controlled, subject-irrelevant regions, often resulting in disproportionate figures such as ``large head with a small body" or ``half body" (see~\cref{failure_case2}). We attribute this to the model’s limited ability to generate subject sizes differing from those in the reference image. To address this, we adopt a gradual training strategy with reference images of increasing size variance, enabling faster convergence and better generalization to varied subject sizes. For $384^2$ reference images, we implement random cropping with progressively larger intervals:

\begin{align}
    \footnotesize
    r_{\text{cur}} = r_{\text{min}} + (r_{\text{max}} - r_{\text{min}}) \times \frac{S_{\text{cur}}}{S_{\text{total}}},
    r_{\text{sample}} \sim U(r_{\text{min}}, r_{\text{cur}}).
\end{align}
Here, $S_{\text{cur}}$ and $S_{\text{total}}$ are the current and total training steps. $r_{\text{min}}$, $r_{\text{max}}$, and $r_{\text{cur}}$ denote the minimum, maximum, and current image resize ratios. At each training step, a random ratio $r_{\text{sample}}$ is drawn from $U(r_{\text{min}}, r_{\text{cur}})$. The reference image is then resized to $(384 \times r_{\text{sample}})^2$, and a $384^2$ patch is randomly cropped as the final reference image for feature extraction. Empirically, we set $r_{\text{min}}=1.0$ and $r_{\text{max}}=\sqrt{10}$, so the cropped patch can be as small as 10\% of the original image area.

\subsection{Inference Framework}
\label{paradigm_inference}

As shown in~\cref{framework}(b), the inference framework of RealCustom++ employs two branches at each generation step: a Guidance Branch, with the visual condition set to \texttt{None}, and a Generation Branch, conditioned on the target subjects. These branches are connected by the \emph{Adaptive Mask Guidance (AMG)}. Given the previous output $\boldsymbol{z_t}$, the Guidance Branch conducts text-conditional denoising to produce a guidance mask, which is subsequently applied in the Generation Branch.

\subsubsection{Guidance Branch}

\textbf{\emph{More spatially accurate image guidance mask: }} On one hand, the textual cross-attention in pre-trained diffusion models primarily captures high-level semantic correspondences, resulting in low-resolution cross-attention maps that are more focused and less noisy, whereas high-resolution cross-attention maps tend to be more diffuse and noisy. On the other hand, self-attention mainly captures low-level, per-pixel correspondences, producing high-resolution self-attention maps that are more fine-grained and spatially accurate. Leveraging these complementary properties, we propose to construct a more spatially precise image guidance mask than our conference version~\cite{huang2024realcustom} by multiplying the low-resolution cross-attention maps of the target native real words (\emph{e.g.}, the real word ``toy'' in ~\cref{framework}) with the high-resolution self-attention maps in the Guidance Branch. Specifically, we first extract all low-resolution cross-attention maps corresponding to the target native real words and resize them to the largest map size (\emph{e.g.}, $64 \times 64$ in Stable Diffusion XL), denoted as $\boldsymbol{M}_{\text{cross}} \in \mathbb{R}^{64 \times 64}$. $\boldsymbol{M}_{\text{cross}}$ is then flattened to a 1D vector, $\boldsymbol{M}_{\text{cross}} \in \mathbb{R}^{4096 \times 1}$. Next, we extract and resize all high-resolution self-attention maps to $\boldsymbol{M}_{\text{self}} \in \mathbb{R}^{4096 \times 4096}$. The final attention map is obtained by element-wise multiplication:
\begin{equation}
    \boldsymbol{M} = \boldsymbol{M}_{\text{self}} \boldsymbol{M}_{\text{cross}} \in \mathbb{R}^{4096 \times 1},
\end{equation}
which is then re-expanded to 2D as $\boldsymbol{M} \in \mathbb{R}^{64 \times 64}$. Next, a Top-K selection is applied: given the target ratio $\gamma_{\text{scope}} \in [0,1]$, only $\lfloor \gamma_{\text{scope}} \times 64 \times 64 \rfloor$ regions with the highest attention scores are retained, while the rest are set to zero. The selected attention map $\boldsymbol{\bar{M}}$ is then normalized by its maximum value as:

\begin{equation}
    \label{max_norm}
    \boldsymbol{\hat{M}} = \frac{\boldsymbol{\bar{M}}}{\text{max}(\boldsymbol{\bar{M})}},
\end{equation}
where $\text{max}(\cdot)$ represents the maximum value. The rationale for this normalization, rather than simply setting it to binary, is that even within these selected parts, their subject relevance varies. Therefore, different regions should have different weights to ensure smooth generation results between masked and unmasked regions.

\begin{figure}
  \centering
  \includegraphics[width=1.0\linewidth]{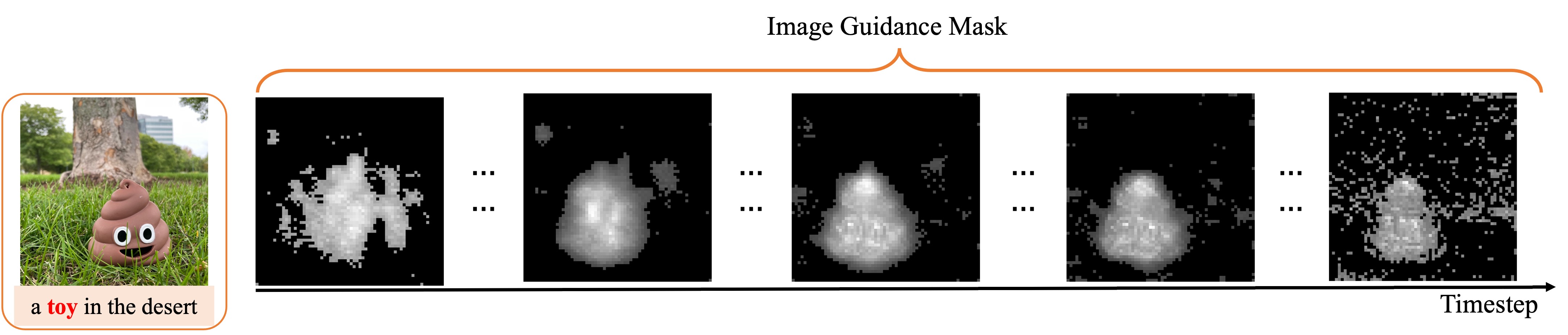}
  \caption{Illustration of the motivation behind the early stop regularization for mask calculation, where the image guidance mask tends to converge in the middle diffusion steps and become more scattered in the later steps.}
  \label{mask_visualization_timestep}
\end{figure}

\textbf{\emph{More temporally stable image guidance mask:}} Through experimentation, we observe that the image guidance mask converges during the middle diffusion steps but becomes increasingly scattered in later steps, as shown in ~\cref{mask_visualization_timestep}. To address this, we introduce an ``early stop'' regularization to stabilize the guidance mask in the later stages. Specifically, given a timestep threshold $T_{\text{stop}}$, we reuse the image guidance mask $\boldsymbol{\hat{M}}_{T_{\text{stop}}}$ for all diffusion steps beyond $T_{\text{stop}}$. This strategy not only stabilizes the guidance mask in the later diffusion steps but also accelerates the customization process, as only a single Generation Branch is required after $T_{\text{stop}}$.

\subsubsection{Generation Branch}

In the Generation Branch, $\boldsymbol{\hat{M}}$ is multiplied with the visual cross-attention results to mitigate any negative impacts on the controllability of the given texts in the subject-irrelevant regions. Specifically, \cref{dual_attention} is reformulated as:

\begin{multline}
    \label{visual_cross_attention_mask}
    \text{Attention}(\boldsymbol{Q},\boldsymbol{K},\boldsymbol{V},\boldsymbol{K_i},\boldsymbol{V_i}) = \\
    \text{Softmax}(\boldsymbol{Q}\boldsymbol{K}^{\top})\boldsymbol{V} + \text{Softmax}(\boldsymbol{Q}\boldsymbol{K_i}^{\top})\boldsymbol{V_i} \boldsymbol{\hat{M}},
\end{multline}
where the necessary resize operation is applied to match the size of $\boldsymbol{\hat{M}}$ with the resolution of each cross-attention block. The denoised output of Generation Branch is denoted as $\boldsymbol{z_{t-1}}$. Classifier-free guidance \cite{ho2022classifier} is applied to produce the next step's denoised latent feature $\boldsymbol{z_{t-1}}$ as:
\begin{equation}
    \boldsymbol{z_{t-1}} = \epsilon_\theta(\emptyset) + \omega ( \boldsymbol{z_{t-1}} - \epsilon_\theta(\emptyset)),
\end{equation}
where $\epsilon_\theta(\emptyset)$ is the unconditional denoised output and $\omega$ is the classifier-free guidance strength.

\subsection{Extension To Multi-subjects Customization}
\label{paradigm_inference_multiple}
Unlike the pseudo word paradigm, which learns subject-specific mappings, RealCustom++ aligns visual conditions with all real words during training, enabling flexible subject customization and  extension to multiple subjects at inference.

\textbf{\emph{Extension to decoupling multiple subjects from a single reference image (One2Many).}} Unlike previous pseudo word paradigm methods that require region-wise attention regularization or pre-defined masks to separate multiple subjects within a single reference, RealCustom++ achieves this without additional design: simply assign different native real words to represent each target subject in the reference image. As shown in ~\cref{introduction_cases}(b), using ``boy'' to customize only the boy with green hair, or ``boy'' and ``horse'' together to customize both subjects, naturally disentangles them due to the robust alignment between visual and textual conditions.

\textbf{\emph{Extension to composing multiple subjects from multiple reference images (Many2Many).}} Given $N$ reference images $\{x^1, x^2, ..., x^N\}$, each representing a distinct subject, we first employ the proposed cross-layer cross-scale projector to encode them into $N$ vision conditions $\{ f_{ci}^{1}, f_{ci}^{2}, ..., f_{ci}^{N} \}$. The visual cross-attention process described in \cref{visual_cross_attention_mask} is then applied $N$ times, once for each vision condition. Here, the Top-K selection for the guidance mask is extended to a novel Multi-Subject Top-K selection, as illustrated in \cref{multiple_subjects}. The core of this algorithm is to iteratively select the highest scores for each subject while ensuring that the resulting guidance masks do not overlap. After obtaining the selected attention map for each subject, $\boldsymbol{\bar{M}}_{j}$ $(j \in [0, N-1])$, maximum normalization is applied to each:

\begin{equation}
    \boldsymbol{\hat{M}}_{j} = \frac{\boldsymbol{\bar{M}}_{j}}{\text{max}(\boldsymbol{\bar{M})}_{j}}, j \in [0, N-1].
\end{equation}
Then the \cref{visual_cross_attention_mask} is rewritten as:

\begin{multline}
    \text{Attention}(\boldsymbol{Q},\boldsymbol{K},\boldsymbol{V},\boldsymbol{K_i},\boldsymbol{V_i}) = 
    \text{Softmax}(\boldsymbol{Q}\boldsymbol{K}^{\top})\boldsymbol{V} \\ + 
    \sum_{j=0}^{N-1} \text{Softmax}(\boldsymbol{Q}\boldsymbol{K_i^j}^{\top})\boldsymbol{V_i^j} \boldsymbol{\hat{M}}_{j},
\end{multline}
where $\boldsymbol{K_i^j}, \boldsymbol{V_i^j}$ stand for the projected vision key and value for $j^{th}$ reference image, respectively.

\begin{algorithm}[!h]
    \small
    \caption{Guidance Mask Construction Algorithm for Multiple Subjects Customization.}
    \label{multiple_subjects}
    \renewcommand{\algorithmicrequire}{\textbf{Input:}}
    \renewcommand{\algorithmicensure}{\textbf{Output:}}
    \begin{algorithmic}[1]
        \REQUIRE given subject number $N$, the multiplied map before selection for each subject $\boldsymbol{M}_{j}, i \in [0, N-1]$, the target ratio for each subject $\gamma_{\text{scope}}^{j}, j \in [0, N-1]$.
        \ENSURE the selected attention map for each subject $\boldsymbol{\bar{M}}_{j}, j \in [0, N-1]$.
    \STATE $\gamma_{\text{num}}^{j} = \lfloor \gamma_{\text{scope}}^{j} \times 64 \times 64 \rfloor$, where $j \in [0, N-1]$;
    \STATE $\gamma_{\text{current}}^{j} = 0$, where $j \in [0, N-1]$;
    \STATE $\boldsymbol{\bar{M}}_{j} = \boldsymbol{0}, j \in [0, N-1]$;
    \STATE $\boldsymbol{M}_{\text{flag}} = \boldsymbol{0}$ 
    \COMMENT{A flag mask that denotes whether a position has been allocated to a given subject, 1 denotes for allocated};
    \WHILE{$ \sum_{j}^{N-1}(\gamma_{\text{current}}^{j} < \gamma_{\text{num}}^{j}$)}

        \FOR{$j = 0$ to $N-1$}
            \IF{$\gamma_{\text{current}}^{j} < \gamma_{\text{num}}^{j}$}
                \STATE \texttt{Set\_NegInf}($\boldsymbol{M}_{j}$, $\boldsymbol{M}_{\text{flag}}$)
                \COMMENT{For positions in $\boldsymbol{M}_{\text{flag}}$ that are 1, the corresponding values in $\boldsymbol{M}_{j}$ are set to negative infinity to avoid repeat selection};
                \STATE (h, w) = \texttt{Copy\_Top1}($\boldsymbol{\bar{M}}_{j}$, $\boldsymbol{M}_{j}$)
                \COMMENT{Copy the maximum value in $\boldsymbol{M}_{j}$ to the corresponding position in $\boldsymbol{\bar{M}}_{j}$, (h, w) is the position of the maximum value};
                \STATE \texttt{Set\_Flag}((h, w), $\boldsymbol{M}_{\text{flag}}$)\COMMENT{set the value of the position (h, w) in $\boldsymbol{M}_{\text{flag}}$ to 1};
                \STATE $\gamma_{\text{current}}^{j} = \gamma_{\text{current}}^{j} + 1$
            \ENDIF
        \ENDFOR
    \ENDWHILE
    \end{algorithmic}
\end{algorithm}

\begin{table*}
    \centering
    \caption{Quantitative comparisons with state-of-the-art methods.}
    \begin{tabular}{ccccccccc}
    \toprule
    \multirow{2}*{Method} & \multirow{2}*{BaseModel} & \multicolumn{3}{c}{\textbf{\emph{text controllability}}} & \multicolumn{3}{c}{\textbf{\emph{subject similarity}}} &  \\
    \cmidrule(lr){3-5} \cmidrule(lr){6-8}
         & & CLIP-B-T$\uparrow$(\%) & CLIP-L-T$\uparrow$(\%) & IR$\uparrow$ & CLIP-B-I$\uparrow$(\%) & CLIP-L-I$\uparrow$(\%) & DINO-I$\uparrow$(\%) \\
    \midrule
    DreamBooth$_{'2023}$ \cite{ruiz2023dreambooth} &  SD-v1.5 & 28.20 & 23.91 & 0.1856 & \underline{84.29} & \underline{83.22} & \underline{71.31} \\
    Custom Diffusion$_{'2023}$ \cite{kumari2023multi} & SD-v1.5 & 28.94 & 25.29 & \underline{0.2401} & 82.78 & 81.54 & 68.42 \\
    DreamMatcher$_{'2024}$\cite{nam2024dreammatcher} & SD-v1.5 & 29.16 & 25.37 & 0.2209 & 83.91 & 82.85 & 69.11 \\
    ELITE$_{'2023}$ \cite{wei2023elite} & SD-v1.5 & 28.72 & 25.07 & -0.0527 & 80.76 & 78.92 & 66.86 \\
    BLIP-Diffusion$_{'2024}$ \cite{li2024blip} & SD-v1.5 & 27.94 & 24.32 & -0.6376 & 82.88 & 80.93 & 67.38 \\
    IP-Adapter$_{'2023}$\cite{ye2023ip} & SD-v1.5 & 26.54 & 22.63 & -0.6199 & 83.20 & 81.56 & 68.00 \\
    Kosmos-G$_{'2024}$\cite{pankosmos} & SD-v1.5 & 25.69 & 21.26 & -0.5177 & 81.94 & 80.20 & 65.24 \\
    MoMA$_{'2024}$ \cite{song2024moma} & SD-v1.5 & \underline{30.81} & \underline{26.75} & 0.1697 & 80.84 & 79.16 & 66.73 \\
    SSR-Encoder$_{'2024}$\cite{zhang2024ssr} & SD-v1.5 & 29.55 & 25.72 & 0.0768 & 81.92 & 79.58 & 67.13 \\
    RealCustom$_{'2024}$\cite{huang2024realcustom} & SD-v1.5 & 31.07 & 27.08 & 0.4871 & 84.62 & 83.51 & 71.49 \\ 
    \textbf{RealCustom++(ours)} & SD-v1.5 & \textbf{31.92} & \textbf{28.72} & \textbf{0.7628} & \textbf{86.36} & \textbf{84.16} & \textbf{73.14} \\
    \hline
    Custom Diffusion$_{'2023}$ \cite{kumari2023multi} & SDXL & 30.87 & 28.49 & \underline{0.6255} & \underline{85.01} & \underline{84.11} & \underline{69.79} \\
    IP-Adapter$_{'2023}$\cite{ye2023ip} & SDXL & 29.74 & 27.32 & 0.1807 & 84.93 & 83.02 & 69.43 \\
    Emu-2$_{'2024}$\cite{sun2024generative} & SDXL & 26.34 & 22.01 & -0.4293 & 83.33 & 82.19 & 67.32 \\
    $\lambda$-ECLIPSE$_{'2024}$\cite{patel2024lambda} & Kandinsky v2.2 & 27.19 & 22.87 & -0.3876 & 84.60 & 83.21 & 68.62 \\
    MS-Diffusion$_{'2024}$ \cite{wang2024ms} & SDXL & \underline{31.24} & \underline{29.03} & 0.5522 & 82.84 & 82.01 & 69.33 \\
    \textbf{RealCustom++(ours)} & SDXL & \textbf{33.43} & \textbf{31.20} & \textbf{1.1036} & \textbf{87.32} & \textbf{86.67} & \textbf{74.60} \\
    \bottomrule
    \end{tabular}
    \begin{tablenotes}
    \footnotesize
    \item RealCustom++ consistently surpasses existing methods on SD-v1.5 and SDXL. It improves text controllability by 7.01\%, 7.48\%, and 76.43\% on CLIP-B-T, CLIP-L-T, and ImageReward, respectively, with the large ImageReward gain highlighting superior disentanglement of text and subject information. For subject similarity, it achieves state-of-the-art results on CLIP-B-I, CLIP-L-I, and DINO-I, demonstrating higher fidelity in subject-relevant regions.
    \end{tablenotes}
    \label{main_result}
\end{table*}

\section{Experiments}

We first describe our experimental setups in \cref{section_experimental_setups}. Then, we compare our proposed RealCustom++ with state-of-the-art customization methods in \cref{section_comparison_with_sota} to demonstrate its superiority in subject similarity and text controllability. Finally, the ablation study of each component of RealCustom++ is presented in \cref{section_ablation}.

\subsection{Experimental Setups}
\label{section_experimental_setups}

\textbf{Implementation Details.} 
RealCustom++ is implemented on SD-1.5~\cite{rombach2022high} and SDXL~\cite{podell2023sdxl}, and trained on the \texttt{<text-image>} dataset (\emph{i.e.}, a filtered subset of LAION-5B~\cite{schuhmann2022laion} based on aesthetic score) and the \texttt{<multi-view>} dataset (\emph{i.e.}, MVImageNet~\cite{yu2023mvimgnet}), as detailed in ~\cref{Curriculum_Training_Recipe}. Training used 8 A100 GPUs for 160{,}000 iterations with a learning rate of $1 \times 10^{-4}$. We employ SigLip~\cite{zhai2023sigmoid} and DINO~\cite{caron2021emerging} as image encoders, both with $384^2$ input size, concatenating their last hidden states along the channel dimension to obtain deep image features in $\mathbb{R}^{729 \times 2176}$, where $n_{\text{image}} = 729$ and $c_{\text{image}} = 2176$. For shallow features, we select $L=3$ layers: \{7, 13, 19\} for SigLip and \{4, 10, 16\} for DINO, concatenated along the channel dimension to yield triple-level shallow features. For SD-1.5, DDIM sampler~\cite{song2020denoising} with 50 steps and 12.5 classifier-free guidance are used. For SDXL, DDIM sampler with 25 steps and 7.5 classifier-free guidance are used. The Top-K ratio $\gamma_{\text{scope}}$ is set to 0.2 by default. The timestep threshold $T_{\text{stop}}$ for early stop mask regularization is set to 25 for SD-v1.5 and 12 for SDXL.

\textbf{Evaluation Metrics.} We comprehensively evaluate RealCustom++ using standard automatic metrics for subject similarity and text controllability. \textbf{(i) \emph{Subject Similarity:}} We use the SAM segmentation model~\cite{kirillov2023segment} to extract subjects and measure similarity with CLIP-I and DINO~\cite{caron2021emerging} scores, calculated as the average pairwise cosine similarity between embeddings of segmented subjects in generated and reference images. For robustness, we report results with both CLIP ViT-B/32 (\texttt{CLIP-B-I}) and CLIP ViT-L/14 (\texttt{CLIP-L-I}), with the latter offering finer-grained assessment. \textbf{(ii) \emph{Text Controllability:}} We compute the cosine similarity between prompt and image embeddings using CLIP ViT-B/32 (\texttt{CLIP-B-T}) and CLIP ViT-L/14 (\texttt{CLIP-L-T}). We also employ ImageReward~\cite{xu2023imagereward} (\texttt{IR}) to jointly assess text controllability and image quality.

\textbf{Evaluation Benchmarks.} Following prior works, we use the prompt ``a photo of [category]" to generate images for similarity evaluation. The full set of editing prompts and subject images is provided in the Appendix. These are based on the standard DreamBench~\cite{ruiz2023dreambooth} and further supplemented with more challenging user-provided cases from open domains, such as cartoon characters and buildings, enabling a more comprehensive evaluation than our conference version. Additionally, we conduct multi-subject experiments on MS-Bench~\cite{wang2024ms} for more comprehensive validation.

\begin{figure*}
  \centering
  \includegraphics[width=0.9\linewidth]{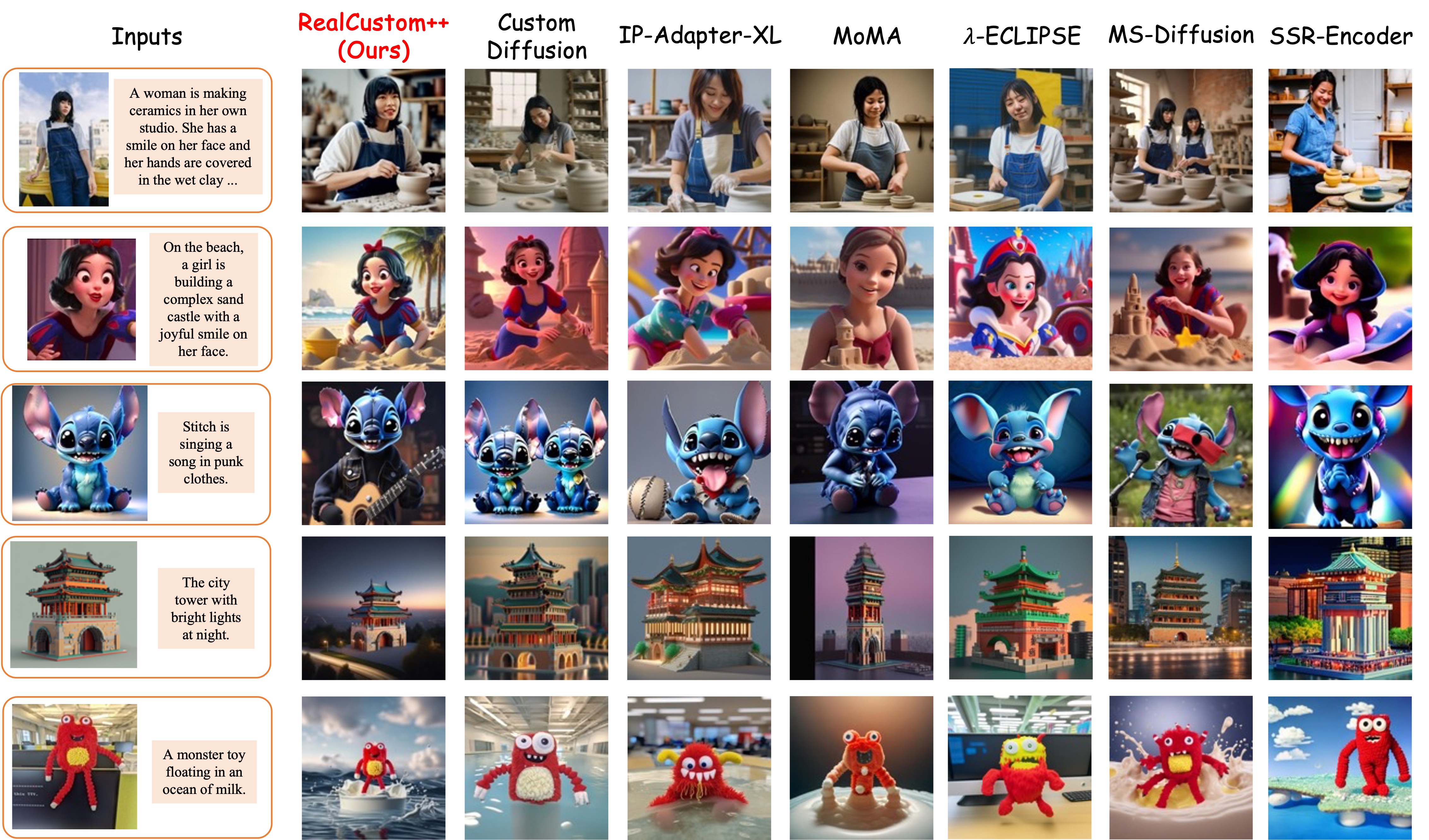}
  \caption{Qualitative results show that RealCustom++ outperforms existing methods in subject fidelity, text alignment, and diversity. It also generates backgrounds that better match the prompts, avoiding artifacts and irrelevant elements, demonstrating superior adaptability and controllability.}
  \label{main_visual_results}
\end{figure*}

\begin{figure*}
  \centering
  \includegraphics[width=0.9\linewidth]{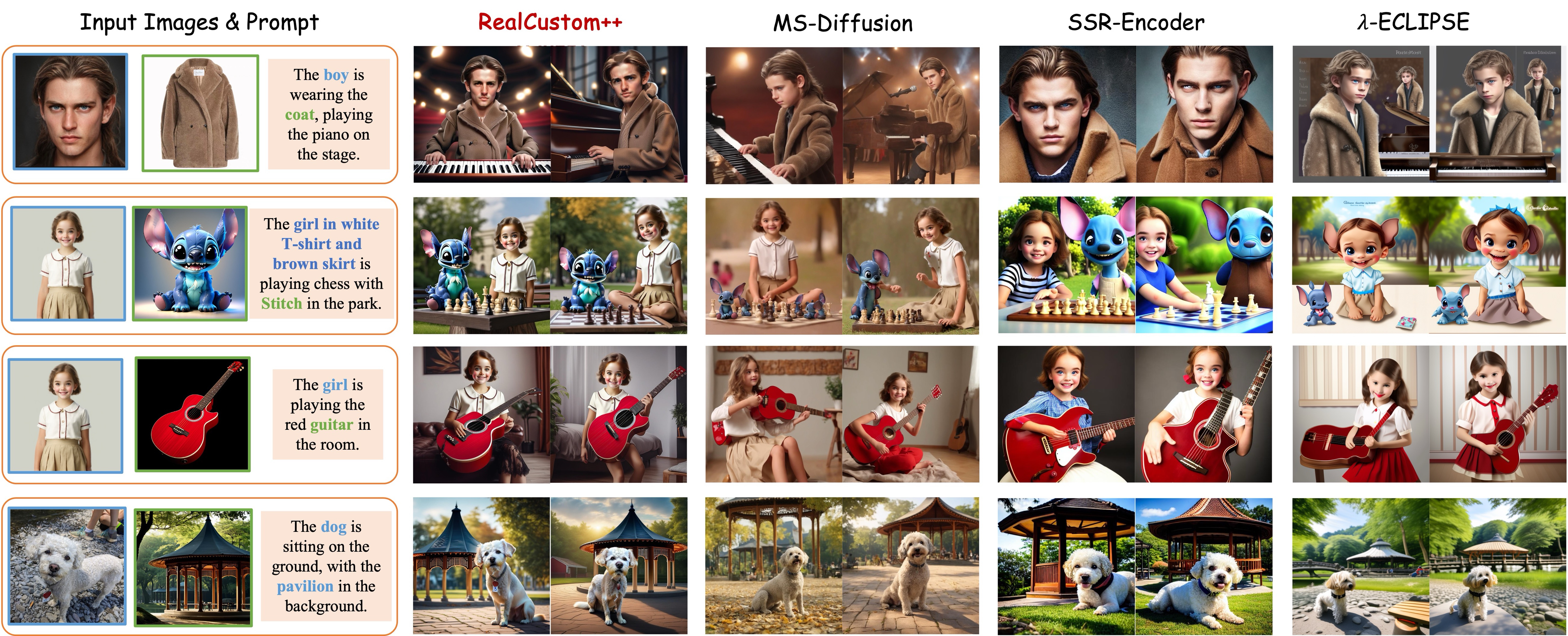}
  \caption{Qualitative comparison for multiple-subject customization. RealCustom++ produces results with superior text controllability and subject similarity compared to existing methods. For instance, in the second row, RealCustom++ successfully generates the scene of the specified ``girl'' playing chess with ``Stitch,'' whereas other methods struggle to maintain consistency across both subjects. Additionally, RealCustom++ effectively handles diverse and complex subject interactions, such as ``wearing'' in the first row and ``playing guitar'' in the fourth row.}
  \label{main_results_multiple_fig}
\end{figure*}

\subsection{Comparison with Stat-of-the-Arts}
\label{section_comparison_with_sota}

\textbf{Quantitative main results.} 
As shown in~\cref{main_result}, RealCustom++ surpasses existing methods on all metrics for both SD-v1.5 and SDXL. (1) For text controllability, RealCustom++ achieves relative improvements of 7.01\%, 7.48\%, and 76.43\% on CLIP-B-T, CLIP-L-T, and ImageReward, respectively, on SDXL. The significant ImageReward gain underscores our method’s effective disentanglement of text and subject, allowing precise control over subject-irrelevant regions. (2) For subject similarity, RealCustom++ sets new state-of-the-art results on CLIP-B-I, CLIP-L-I, and DINO-I, demonstrating superior fidelity in subject-relevant areas. (3) Compared to our conference version~\cite{huang2024realcustom}, RealCustom++ further improves text controllability (CLIP-L-T) by 6.06\%, subject similarity (DINO-I) by 2.31\%, and overall image quality by 56.6\%.

\textbf{Qualitative main results.} As shown in \cref{main_visual_results}, RealCustom++ achieves superior zero-shot open-domain customization across diverse subjects, including humans, characters, buildings, animals, and uniquely shaped toys. It consistently delivers higher-quality images with enhanced subject similarity and text controllability compared to existing methods. The effective disentanglement of subject and text further enables RealCustom++ to generate cleaner, more contextually appropriate backgrounds; for instance, in the second row, it accurately renders the ``on the beach" prompt, unlike prior methods that retain irrelevant background elements.

\textbf{Multiple subjects customization.} To evaluate multiple-subject customization, we follow previous works~\cite{liu2023cones, wang2024ms} and collect 30 cases spanning animals, characters, buildings, and objects. Fully customized subjects and prompts are provided in the Appendix. Subject similarity is measured as the mean similarity across all subjects. As shown in~\cref{multi_subject_result}, RealCustom++ outperforms state-of-the-art methods, achieving a 4.6\% improvement on CLIP-B-T for text controllability, and 6.34\% and 3.9\% improvements on CLIP-B-I and DINO-I for subject similarity. Qualitative results in~\cref{main_results_multiple_fig} further confirm that RealCustom++ delivers superior controllability and similarity. For example, in the second row, RealCustom++ accurately generates the scene of the specified ``girl" playing chess with ``Stitch", while other methods fail to maintain consistency for both subjects. RealCustom++ also handles complex subject interactions, such as ``wearing" in the first row and ``playing guitar" in the fourth row. These results further validate the effectiveness of the RealCustom++.

\begin{table}
    \centering
    \caption{Multiple-Subject Customization Comparisons}
    \begin{tabular}{ccccccccc}
    \toprule
    Methods & CLIP-B-T(\%) & CLIP-B-I(\%) & DINO-I(\%) \\
    \hline
    CustomDiffusion \cite{kumari2023multi} & 27.24 & \underline{80.23} & \underline{63.78} \\
    $\lambda$-ECLIPSE\cite{patel2024lambda} & 28.05 & 76.32 & 59.23 \\
    SSR-Encoder\cite{zhang2024ssr} & 28.43 & 79.25 & 62.08 \\
    MS-Diffusion\cite{wang2024ms} & \underline{30.02} & 78.25 & 61.28 \\
    \textbf{RealCustom++(ours)} & \textbf{31.4} & 
    \textbf{85.32} & \textbf{66.29} \\
    \bottomrule
    \end{tabular}
    \begin{tablenotes}
    \footnotesize
    \item RealCustom++ achieves state-of-the-art performance, with a 4.6\% improvement on CLIP-B-T for text controllability, and 6.34\% and 3.9\% improvements on CLIP-B-I and DINO-I for subject similarity.
    \end{tablenotes}
    \label{multi_subject_result}
\end{table}

\begin{table}
    \centering
    \caption{Multiple-Subject Customization on MS-Bench}
    \begin{tabular}{ccccccccc}
    \toprule
    Methods & CLIP-I & DINO & M-DINO & CLIP-T \\
    \hline
    $\lambda$-ECLIPSE\cite{patel2024lambda} & 0.724 & 0.419 & 0.094 & 0.316 \\
    SSR-Encoder\cite{zhang2024ssr} & \underline{0.725} & \underline{0.425} & 0.107 & 0.303 \\
    MS-Diffusion\cite{wang2024ms} & 0.698 & \underline{0.425} & \underline{0.108} & \underline{0.341} \\
    \textbf{RealCustom++(ours)} & \textbf{0.741} & \textbf{0.440} & \textbf{0.214} & \textbf{0.347} \\
    \bottomrule
    \end{tabular}
    \label{multi_subject_result_ms_bench}
\end{table}

\textbf{Multiple subjects customization on MS-Bench.} To provide a more comprehensive evaluation, we further test on MS-Bench~\cite{wang2024ms}, which contains 40 subjects, primarily clothes and objects from DreamBench. Following the protocol of MS-Diffusion~\cite{wang2024ms}, results are shown in \cref{multi_subject_result_ms_bench}. RealCustom++ achieves the best performance across all metrics, particularly on M-DINO, which assesses whether each subject is faithfully recreated. These results highlight the effectiveness of our proposed multiple adaptive mask guidance, which provides accurate and non-overlapping guidance masks for each subject, thereby enabling faithful generation of each subject.

\subsection{Ablations}
\label{section_ablation}

\begin{figure*}
  \centering
  \includegraphics[width=0.85\linewidth]{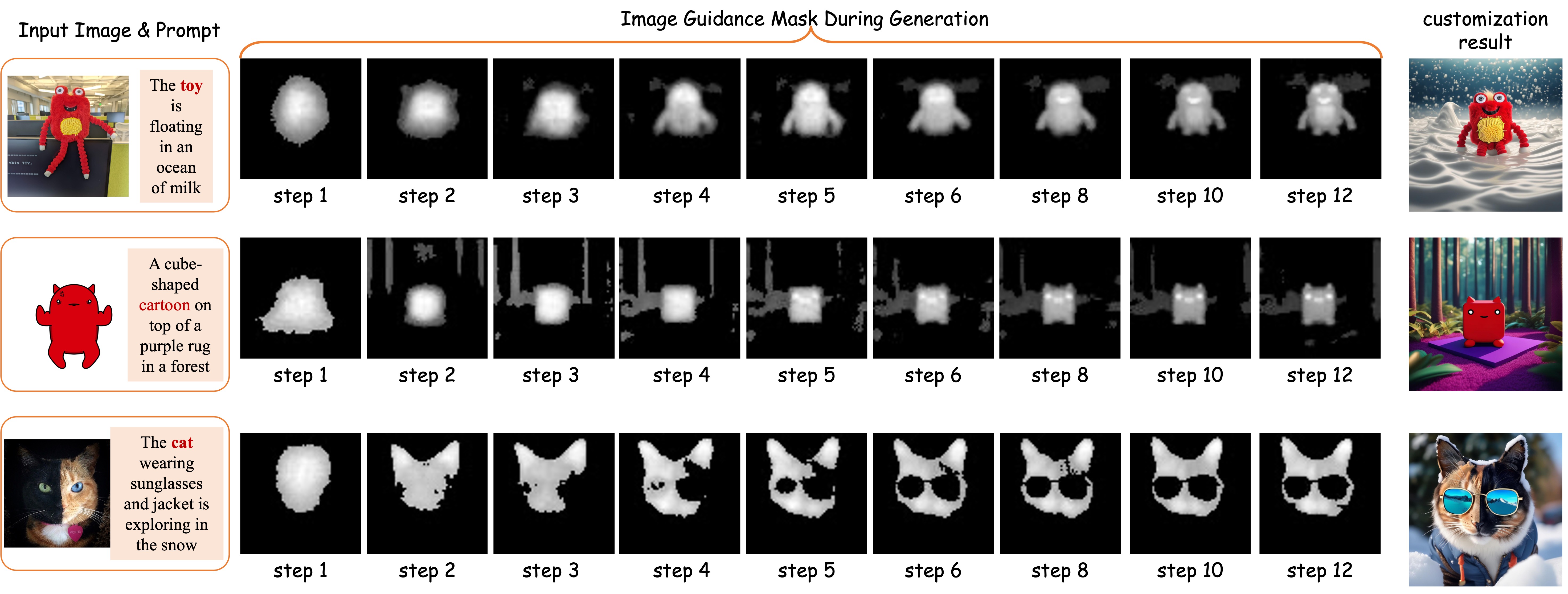}
  \caption{Illustration of the progressive customization of target real words into the given subjects for single-subject customization. Customized words are highlighted in red, with their attention maps gradually forming the target subjects and incrementally adding details. This process yields a more precise image guidance mask for open-domain customization, while subject-irrelevant regions remain fully controlled by the input text.}
  \label{mask_visualization_single}
\end{figure*}

\begin{figure*}
  \centering
  \includegraphics[width=0.85\linewidth]{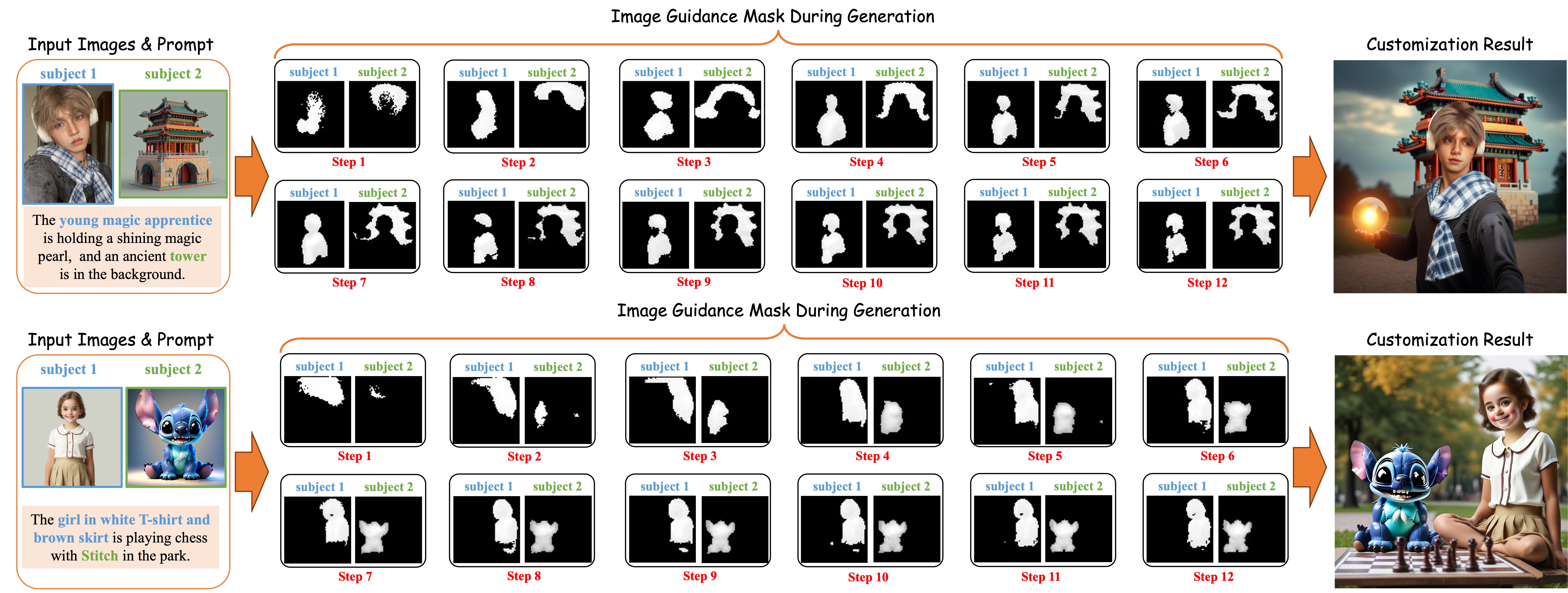}
  \caption{Illustration of progressive customization for multiple subjects: customized words for each subject are highlighted in blue and green. RealCustom++ generates accurate, decoupled guidance masks for each subject, enabling high-quality similarity across all subjects.}
  \label{mask_visualization_multiple}
\end{figure*}

\textbf{Effectiveness of Adaptive Mask Guidance (AMG).} 
We visualize the customization process for both single-subject (\cref{mask_visualization_single}) and multiple-subject (\cref{mask_visualization_multiple}) cases. In \cref{mask_visualization_single}, the attention maps for target real words progressively align with the given subjects, adding detail step by step and enabling open-domain zero-shot customization while maintaining full text control over subject-irrelevant regions. RealCustom++ also adapts to shape variations (\emph{e.g.}, second row) and subject overlaps (\emph{e.g.}, third row). In \cref{mask_visualization_multiple}, RealCustom++ generates accurate, decoupled guidance masks for each subject, ensuring high-quality similarity across all subjects.

\begin{table}
    \centering
    \caption{Ablation study on different image mask guidance}
    \begin{tabular}{c l ccccccc}
    \toprule
    No. & Attention Setting & CLIP-B-T $\uparrow$ & CLIP-B-I $\uparrow$ & \\
    \hline
    No.1 & all resolution cross-attn & 32.06 & 83.97 \\
    \hline
    No.2 & \makecell[l]{low resolution cross-attn + \\ (conference version)} & 32.20 & 84.23 \\
    \hline
    No.3 & \makecell[l]{(2) + \\ all resolution self-attn} & 31.08 & 85.22 \\
    \hline
    No.4 & \makecell[l]{(2) + \\ high resolution self-attn \\ (Adaptive Mask Guidance)} & \textbf{33.43} & \textbf{87.32} \\
    \bottomrule
    \end{tabular}
    \begin{tablenotes}
    \footnotesize
    \item The combination of low-resolution cross-attention and high-resolution self-attention yields the most accurate image mask for customization. Compared to our conference version (No.2), the new guidance mask (No.4) achieves improvements of 3.8\% in controllability (CLIP-B-T) and 3.7\% in similarity (CLIP-B-I) simultaneously.  
    \end{tablenotes}
    \label{ablation_mask_guidance_1}
\end{table}

\begin{figure*}
  \centering
  \includegraphics[width=0.85\linewidth]{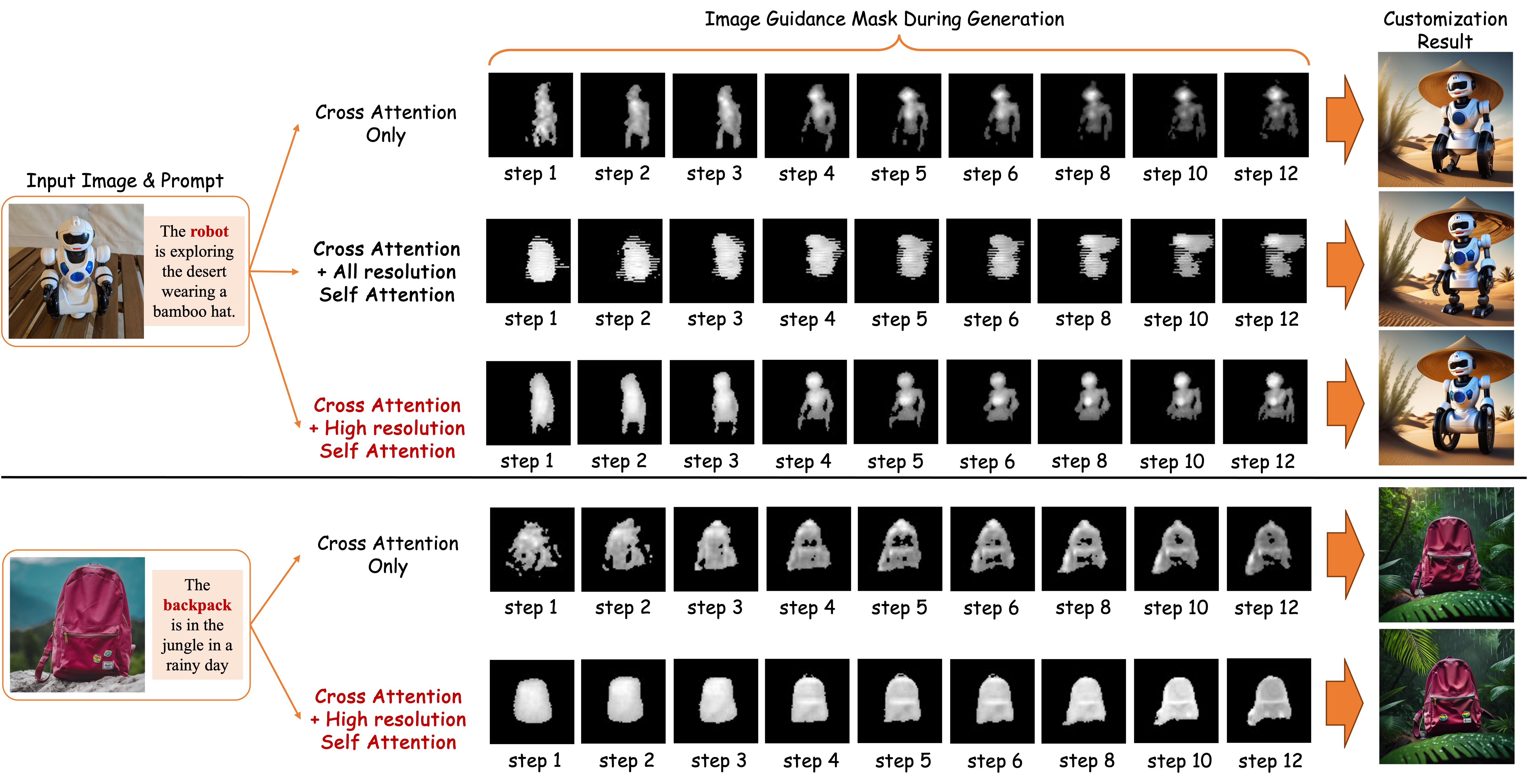}
  \caption{Visualization of using different attention mask to construct image guidance mask. We show that (1) solely relying on cross attention map results in scattered mask guidance results in the degradation of subject details; (2) using all resolution self-attention map will make the guidance mask over-focused, resulting in the degradation of mask accuracy.}
  \label{mask_visualization_selfattn}
\end{figure*}

We ablate different attention mechanisms for image mask guidance. As shown in \cref{ablation_mask_guidance_1}, combining low-resolution cross-attention and high-resolution self-attention produces the most accurate masks, optimizing both text controllability and subject similarity. Visualizations in \cref{mask_visualization_selfattn} show that using only cross-attention leads to incomplete masks, while only self-attention results in over-focused, less accurate masks.

\begin{table}
    \centering
    \caption{Ablation study of using different Top-K $\gamma_{\text{scope}}$ ratios}
    \begin{tabular}{lccccccc}
    \toprule
    TopK ratio & CLIP-B-T $\uparrow$ & CLIP-B-I $\uparrow$ & \\
    \hline
    $\gamma_{\text{scope}} = 0.1$ & 33.43 & 84.43 \\
    $\gamma_{\text{scope}} = 0.15$ & 33.43 & 86.09 \\
    $\gamma_{\text{scope}} = 0.2$ & \textbf{33.43} & \textbf{87.32} \\
    $\gamma_{\text{scope}} = 0.2$  w/o mask norm & 30.26 & 87.32 \\
    $\gamma_{\text{scope}} = 0.3$ & 33.29 & 87.32 \\
    $\gamma_{\text{scope}} = 0.4$ & 32.62 & 87.32 \\
    $\gamma_{\text{scope}} = 0.5$ & 31.08 & 87.32 \\
    \bottomrule
    \end{tabular}
    \label{ablation_mask_guidance_2}
\end{table}

The ablation study on early stop regularization for temporally stable image mask guidance is shown in \cref{ablation_mask_guidance_3}. This supports our observation in \cref{mask_visualization_timestep} that guidance masks converge at intermediate diffusion steps, and reusing the mask from these steps enhances both similarity and controllability.

Qualitative results for different Top-K ratios $\gamma_{\text{scope}}$ are shown in \cref{ablation_mask_guidance_2}. We observe that (1) results are robust within a suitable range ($\gamma_{\text{scope}} \in [0.15, 0.4]$); (2) maximum normalization (\cref{max_norm}) is crucial for balancing similarity and controllability, as it assigns appropriate weights to regions with varying subject relevance; and (3) setting $\gamma_{\text{scope}}$ too low or too high degrades similarity or controllability, respectively.

\begin{table}
    \centering
    \caption{Ablation study of mask calculation early stop regularization}
    \begin{tabular}{lccccccc}
    \toprule
    Early Stop Step & CLIP-B-T $\uparrow$ & CLIP-B-I $\uparrow$ & \\
    \hline
    5 & 32.06 & 85.13 \\
    10 & 33.20 & 87.21 \\
    12 & \textbf{33.43} & \textbf{87.32} \\
    15 & 33.40 & 87.07 \\
    20 & 33.41 & 86.84\\
    \texttt{None} (w/o early stop regularization) & 33.41 & 86.67 \\
    \bottomrule
    \end{tabular}
    \begin{tablenotes}
    \footnotesize
    \item Compared to our conference version (without early stop regularization), the new guidance mask algorithm improves similarity from 86.67 to 87.32.
    \end{tablenotes}
    \label{ablation_mask_guidance_3}
\end{table}

\begin{table}
    \centering
    \caption{Ablation study of the cross-layer cross-scale projector}
    \begin{tabular}{c l ccccccc}
    \toprule
    No. & Projector Setting & CLIP-B-T $\uparrow$ & CLIP-B-I $\uparrow$ & \\
    \hline
    (1) & \scriptsize{MLP(naive projector)} & 33.21 & 83.32 \\
    (2) & \scriptsize{(1)+cross-layer mechanism} & 33.45 & 86.12 \\
    (3) & \scriptsize{\makecell[l]{(2)+cross-scale mechanism \\ (\scriptsize{CCP module})}} & 33.43 & 87.32 \\
    \bottomrule
    \end{tabular}
    \begin{tablenotes}
    \footnotesize
    \item The Cross-layer Cross-scale Projector (CCP) module is designed to improve subject similarity (measured by CLIP-B-I) by adaptively fusing multi-layer and multi-scale subject image features, while ensuring that text controllability (measured by CLIP-B-T) is not compromised.
    \end{tablenotes}
    \label{ablation_projector}
\end{table}

\begin{table}
    \centering
    \caption{Ablation study of different feature combination}
    \begin{tabular}{c c ccccccc}
    \toprule
    Feature Combination Setting & CLIP-B-T $\uparrow$ & CLIP-B-I $\uparrow$ & \\
    \hline
    \scriptsize{\makecell[c]{cross-scale element add \\ + cross-layer token concat}} & 29.94 & 86.33 \\
    \hline
    \scriptsize{\makecell[c]{cross-scale token concat \\ + cross-layer token concat}} & 33.65 & 85.62 \\
    \hline
    \scriptsize{\makecell[c]{cross-scale token concat \\ + cross-layer element add \\ (CCP module)}} & 33.43 & 87.32 \\
    \bottomrule
    \end{tabular}
    \label{ablation_combination}
\end{table}

\begin{table}
    \centering
    \caption{Ablation study of the curriculum training recipe}
    \begin{tabular}{c l ccccccc}
    \toprule
    No. & Training Recipe & CLIP-B-T $\uparrow$ & CLIP-B-I $\uparrow$ & \\
    \hline
    No.1 & \texttt{<text-image>} data only & 30.02 & 86.24 \\
    \hline
    No.2 & \texttt{<multiview>} data only & 32.18 & 83.92 \\
    \hline
    No.3 & \makecell[l]{\texttt{<text-image>} data + \\ \texttt{<multiview>} data + \\ $p^{\text{text-image}} = p^{\text{multiview}} = 0.5 $ } & 31.89 & 85.33 \\
    \hline
    No.4 & \makecell[l]{\texttt{<text-image>} data + \\ \texttt{<multiview>} data + \\ curriculum data strategy} & 32.10 & 86.11 \\
    \hline
    No.5 & \makecell[l]{No.4 + cropping strategy \\ with a fixed random ratio \\ between [1, $\sqrt{10}$]} & 33.43 & 86.82 \\
    \hline
    No.6 & \makecell[l]{No.4 +  \\ curriculum cropping strategy \\ (\scriptsize{Curriculum Training Recipe})} & \textbf{33.43} & \textbf{87.32} \\
    \bottomrule
    \end{tabular}
    \begin{tablenotes}
    \footnotesize
    \item The Curriculum Training Recipe improves both similarity (CLIP-B-I) and controllability (CLIP-B-T). Using both text-image and multiview data boosts both metrics (No.4 \emph{vs.} No.1, No.2), and the curriculum data outperforms random one (No.4 \emph{vs.} No.3). Curriculum cropping further enhances both metrics (No.6 \emph{vs.} No.4), yielding higher similarity than random cropping (No.6 \emph{vs.} No.5) while maintaining controllability.
    \end{tablenotes}
    \label{ablation_training_Recipe}
\end{table}

\textbf{Effectiveness of the Cross-layer Cross-scale Projector (CCP).} As shown in \cref{ablation_projector}, ablation of the cross-layer cross-scale projector demonstrates that both cross-layer and cross-scale enhancements significantly improve subject similarity with minimal impact on text controllability. This supports our design principle of preserving the dominance of deep features to maintain effective alignment with text conditions. Further ablation in \cref{ablation_combination} confirms that element-wise addition is optimal for same-level features, while token-wise concatenation is preferable for different-level features to prevent conflicts.

\textbf{Effectiveness of Curriculum Training Recipe (CTR).} As shown in \cref{ablation_training_Recipe}, our findings are: (1) Training exclusively on \texttt{<text-image>} or \texttt{<multiview>} data reduces either text controllability or subject similarity, due to limited pose or subject diversity, respectively. (2) The curriculum data strategy outperforms simple 50\% mixing, yielding higher CLIP-B-T and CLIP-B-I scores. (3) Incorporating curriculum cropping further boosts both metrics, underscoring its role in achieving stable convergence and improved size generalization.

\textbf{Component-wise Ablation.} We clarify that RealCustom++ adopts a train-inference decoupled framework, in which the Cross-layer Cross-scale Projector (CCP) and Curriculum Training Recipe (CTR) are applied during training, while the Adaptive Mask Guidance (AMG) is applied during inference. Therefore, the effectiveness of the training components (\emph{i.e.}, CCP and CTR) is orthogonal to that of the inference component (\emph{i.e.}, AMG), and we investigate the interactions between training and inference components separately.


\begin{table}[h]
\centering
\caption{Training Component-wise Ablation}
\begin{tabular}{r r l l}
\toprule
\multicolumn{2}{c}{Experiment Setting} & CLIP-B-T$\uparrow$ & CLIP-B-I$\uparrow$ \\
\midrule
\multirow{2}{*}{w/o CCP} & w/o CTR & 29.86 & 83.03 \\
& w/ CTR & 33.21({$\Delta$3.35$\uparrow$}) & 83.32({$\Delta$0.29$\uparrow$}) \\
\hdashline
\multirow{2}{*}{w/ CCP} & w/o CTR & 30.02 & 86.24 \\
& w/ CTR & 33.43({$\Delta$3.41$\uparrow$}) & 87.32({$\Delta$1.08$\uparrow$}) \\
\midrule
\multirow{2}{*}{w/o CTR} & w/o CCP & 29.86 & 83.03 \\
& w/ CCP & 30.02({$\Delta$0.16$\uparrow$}) & 86.24({$\Delta$3.21$\uparrow$}) \\
\hdashline
\multirow{2}{*}{w CTR} & w/o CCP & 33.21 & 83.32 \\
& w/ CCP & 33.43({$\Delta$0.22$\uparrow$}) & 87.32({$\Delta$4.0$\uparrow$}) \\
\bottomrule
\end{tabular}
\begin{tablenotes}
    \footnotesize
    \item Ablation on the interaction between training components, \emph{i.e.}, CCP and CTR, where we evaluate the effect of CTR both with and without the CCP, and vice versa. ``w/'' and ``w/o'' denote for ``with'' and ``without''.
\end{tablenotes}
\label{train_component_wise_ablation}
\end{table}


\begin{table}[h]
\centering
\caption{Inference Component-wise Ablation}
\begin{tabular}{r r l l}
\toprule
\multicolumn{2}{c}{Experiment Setting} & CLIP-B-T$\uparrow$ & CLIP-B-I$\uparrow$ \\
\midrule
\multirow{2}{*}{w/o AMG-S} & w/o AMG-T & 32.30 & 83.65 \\
& w AMG-T & 32.30($\Delta${0}) & 84.23({$\Delta$0.58$\uparrow$}) \\
\hdashline
\multirow{2}{*}{w AMG-S} & w/o AMG-T & 33.41 & 86.67 \\
& w/ AMG-T & 33.43($\Delta${0.02$\uparrow$}) & 87.32($\Delta${0.65$\uparrow$}) \\
\hline
\multirow{2}{*}{w/o AMG-T} & w/o AMG-S & 32.30 & 83.65 \\
& w/ AMG-S & 33.41($\Delta${1.11$\uparrow$}) & 86.67($\Delta${3.02$\uparrow$}) \\
\hdashline
\multirow{2}{*}{w/ AMG-T} & w/o AMG-S & 32.30 & 84.23 \\
& w/ AMG-S & 33.43($\Delta${1.13$\uparrow$}) & 87.32($\Delta${3.09$\uparrow$}) \\
\bottomrule
\end{tabular}
\begin{tablenotes}
    \footnotesize
    \item Ablation on the interaction between inference components, \emph{i.e.}, the self-attention augmentation for spatial mask accuracy (AMG-S) and the early-stop regularization for temporal mask robustness (AMG-T).
\end{tablenotes}
\label{inference_component_wise_ablation}
\end{table}

The interaction between CCP and CTR is shown in \cref{train_component_wise_ablation}, we observe that: (1) \textbf{CCP amplifies CTR’s performance gains on subject similarity (CLIP-B-I)}. Specifically, when CCP is incorporated, CTR improves the CLIP-B-I score by 1.08, compared to only 0.29 without CCP. This enhancement could be attributed to CCP’s ability to provide more robust and fine-grained subject representations, which allow CTR to inject these representations more effectively during training. (2) \textbf{CCP’s performance gains are consistent with or without CTR}. The reason is that CCP is designed to enhance subject representation and acts as a prerequisite for CTR, making its benefits independent of CTR’s presence.


\Cref{inference_component_wise_ablation} presents the component-wise ablation of AMG during inference, including the self-attention augmentation for enhanced spatial mask accuracy (AMG-S), and the early-stop regularization that reuses masks generated in previous steps to enhance mask temporal robustness and accelerate inference (AMG-T). The results show that: (1) \textbf{AMG-T’s performance gains are consistent with or without AMG-S, and vice versa}, demonstrating that they function independently without interference. (2) AMG-S is important for accurate mask guidance throughout the entire inference process, improving controllability and similarity. (3) \textbf{AMG retains robustness without AMG-T}, with only a minor decrease in similarity, as AMG-T primarily affects the final inference steps.

\textbf{Time Analysis.} We clarify that ``real time" in our title indicates that RealCustom++ does not require per-subject finetuning, as is necessary in previous methods like DreamBooth~\cite{ruiz2023dreambooth}, which can take minutes to hours per image. Instead, our method operates with a time cost comparable to standard text-to-image generation as shown in \cref{ablation_time_analysis} (20.45 seconds for RealCustom++ \emph{vs.} 14.21 seconds for standard text-to-image generation). The increased time cost compared to standard text-to-image generation arises from our dual-branch inference, which requires two U-Net forward passes per generation step. With mask calculation early stop regularization, where masks are reused in later steps, the time cost is reduced to about 1.5 times that of standard text-to-image generation. We would like to point out that this slight increase in time cost compared to standard text-to-image generation has minimal impact on practical usage.



\begin{table}
    \centering
    \caption{Time Analysis on SDXL}
    \begin{tabular}{l c}
    \toprule
    Setting & Time \\
    \hline
    Textual Inversion \cite{gal2022image} & $\sim$ 50 min \\
    DreamBooth \cite{ruiz2023dreambooth} & $\sim$ 15 min \\
    Custom Diffusion \cite{kumari2023multi} & $\sim$ 6 min \\
    \hline
    Text-to-Image Generation (SDXL) & 14.21 second \\
    RealCustom++ w/o early stop regularization & 24.09 second \\
    RealCustom++ & 20.45 second \\
    \bottomrule
    \end{tabular}
    \label{ablation_time_analysis}
\end{table}

\begin{figure}
  \centering
  \includegraphics[width=0.9\linewidth]{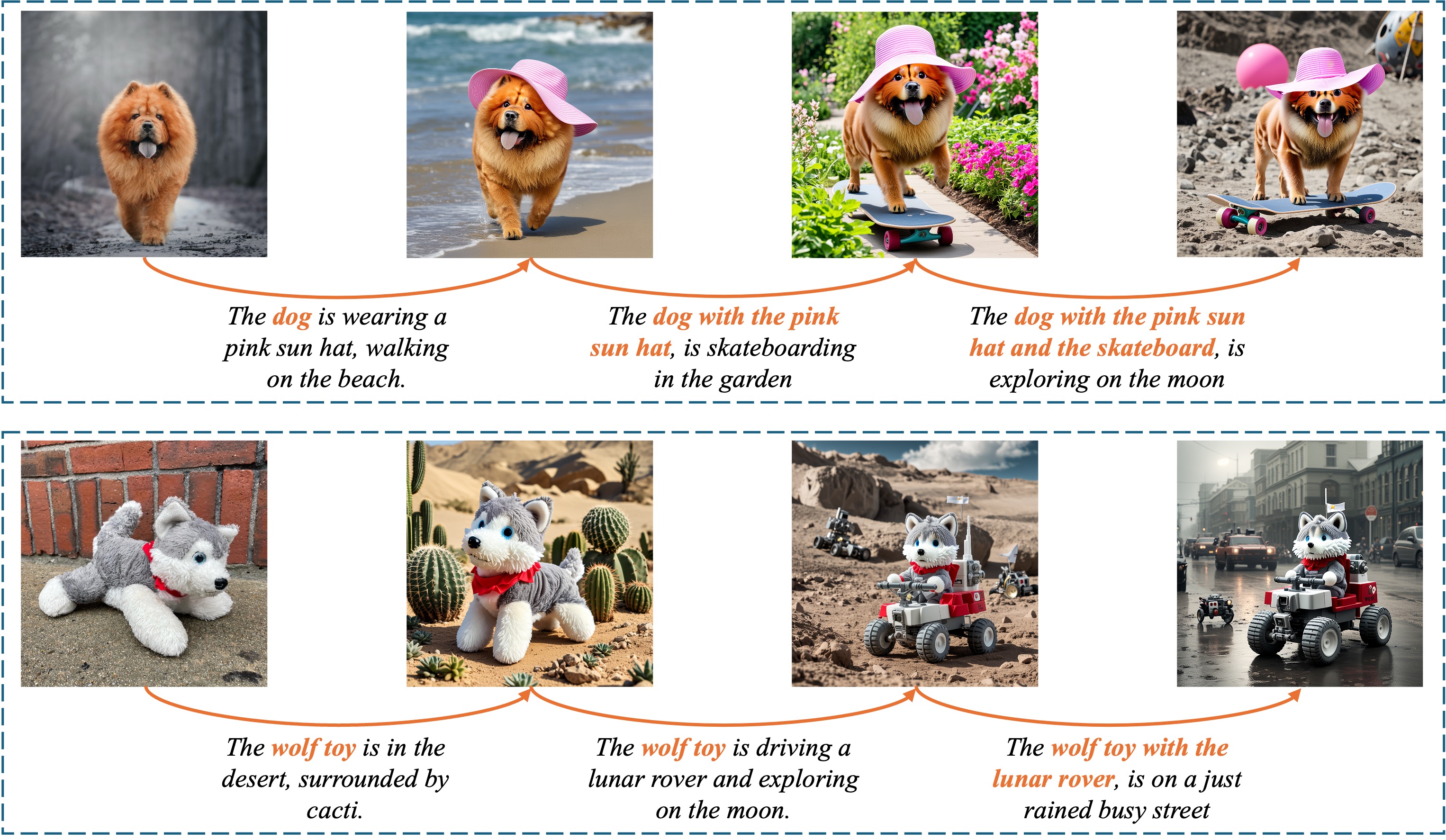}
  \caption{Multi-round generation with RealCustom++. Customized real words for each round are highlighted in orange. RealCustom++ enables flexible multi-round generation by specifying different target real words.}
  \label{multi_round_generation}
\end{figure}

\textbf{Multi-round Generation.} As shown in \cref{multi_round_generation}, our real-word paradigm naturally supports multi-round generation, where the output from each round serves as the reference subject image for the next. This enables flexible customization in each round by specifying different target real words. For example, in the first row, the initial round uses ``dog" as the target word, preserving only the dog's characteristics. In the second round, the target word ``dog with the pink hat" incorporates the pink hat generated in the previous round, allowing RealCustom++ to retain both features. This demonstrates the strong generalization capability of RealCustom++, enabling the progressive accumulation and preservation of subject characteristics across multiple rounds.

\begin{figure}
  \centering
  \includegraphics[width=0.9\linewidth]{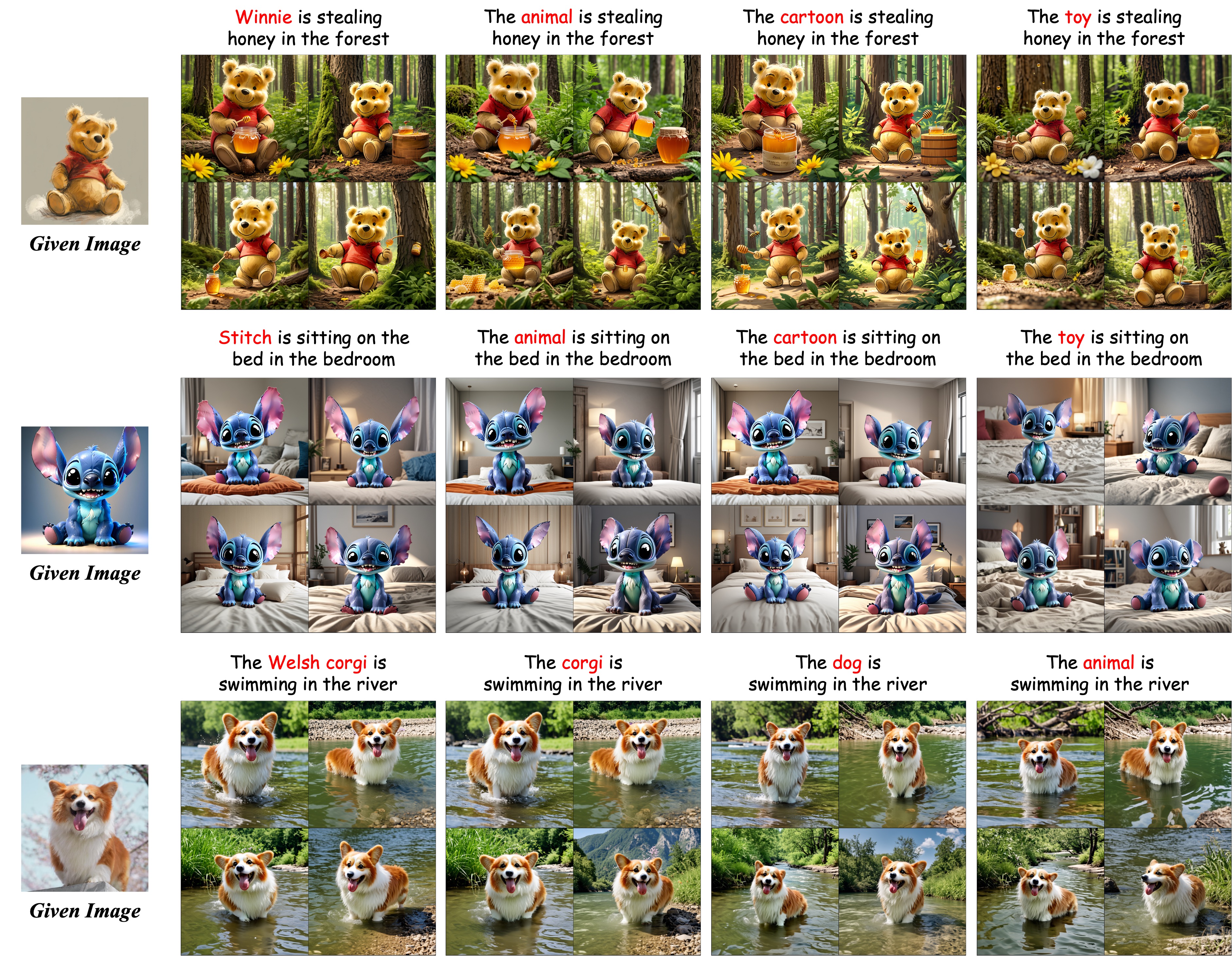}
  \caption{The customization results of using different real words.}
  \label{different_words}
\end{figure}


\textbf{Generalization and Robustness to Different Real Words.} As shown in \cref{different_words}, RealCustom++ generates robust customization results regardless of the granularity of the real words used, from coarse-grained super-category (\emph{e.g.}, animal) to fine-grained specific words (\emph{e.g.}, Winnie, Welsh corgi). It consistently preserves the unique identity of each subject while maintaining alignment with text semantics. This robustness stems from training on a large, generic text-image dataset~\cite{schuhmann2022laion}, which enables RealCustom++ to learn a general alignment between visual conditions and all real words, including both super-category and specific subject words.
\section{Conclusion}
In this paper, we introduce RealCustom++, a novel customization paradigm that, for the first time, represents subjects as non-conflicting real words, enabling precise disentanglement of subject similarity and text controllability. This is realized through a progressive customization process within a train-inference decoupled framework, refining target real words from general concepts to specific subjects. RealCustom++ leverages a cross-layer cross-scale projector and a curriculum training strategy to achieve robust feature extraction and diversity in pose and size. At inference, adaptive mask guidance ensures accurate customization of target real words while preserving subject-irrelevant regions. We further extend RealCustom++ to multi-subject scenarios with a multi-real-word customization algorithm. Extensive experiments demonstrate state-of-the-art performance in subject similarity and text controllability for both single- and multi-subject real-time open-domain customization.

{
\bibliographystyle{IEEEtran}
\bibliography{main}
}


 

\begin{IEEEbiography}
[{\includegraphics[width=1in,height=1.25in,clip,keepaspectratio]{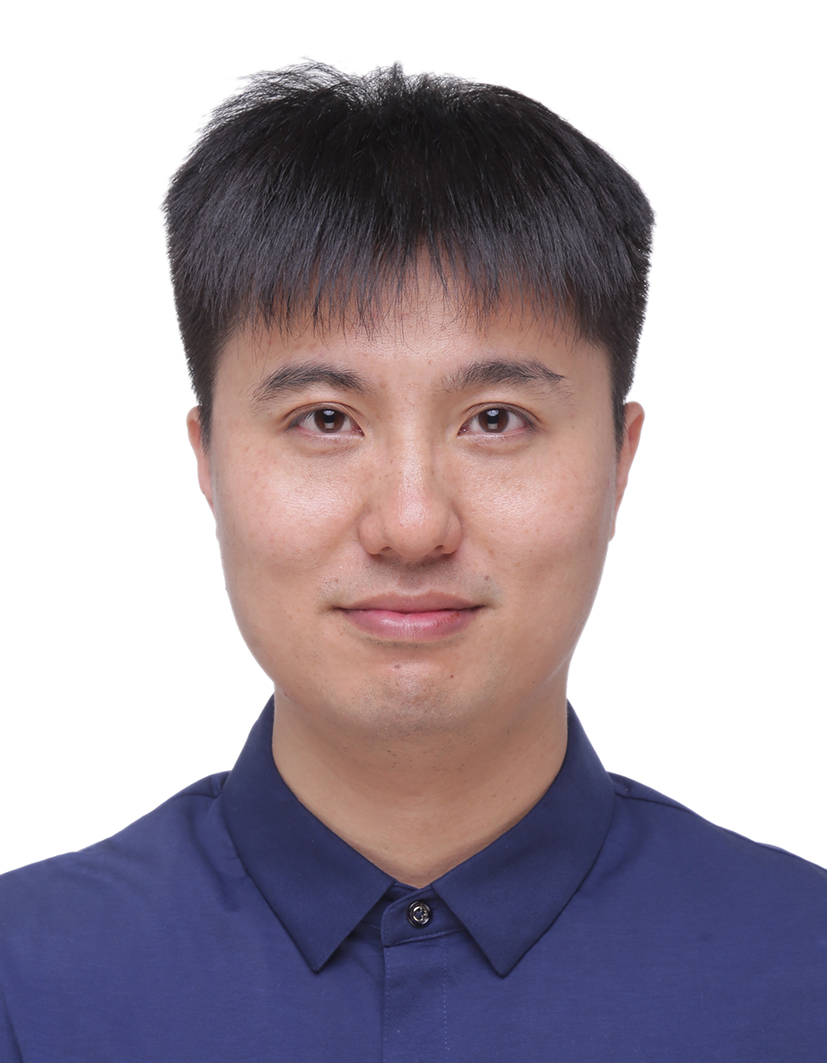}}] {Zhendong Mao} received the Ph.D. degree in computer application technology from the Institute of Computing Technology, Chinese Academy of Sciences, in 2014. He is currently a professor with Department of Electronic Engineering and Information Science, University of Science and Technology of China, Hefei, China. He was an assistant professor with the Institute of Information Engineering, Chinese Academy of Sciences, Beijing, from 2014 to 2018. He has authored more than 70 refereed journal and conference papers, including TPAMI, TKDE, TIP, CVPR and ACL, accumulating more than 7200 citations on Google Scholar. He was a recipient of the best paper award in PCM 2013 and the best student paper award in ACM Multimedia 2022. His research interests include cross-modal understanding and cross-modal generation. He serves as an Associate Editor of the \textit{IEEE 
Transactions on Circuits and Systems for Video Technology (T-CSVT)} and \textit{IEEE Transactions on Multimedia (T-MM)}.
\end{IEEEbiography}

\begin{IEEEbiography}
[{\includegraphics[width=1in,height=1.25in,clip,keepaspectratio]{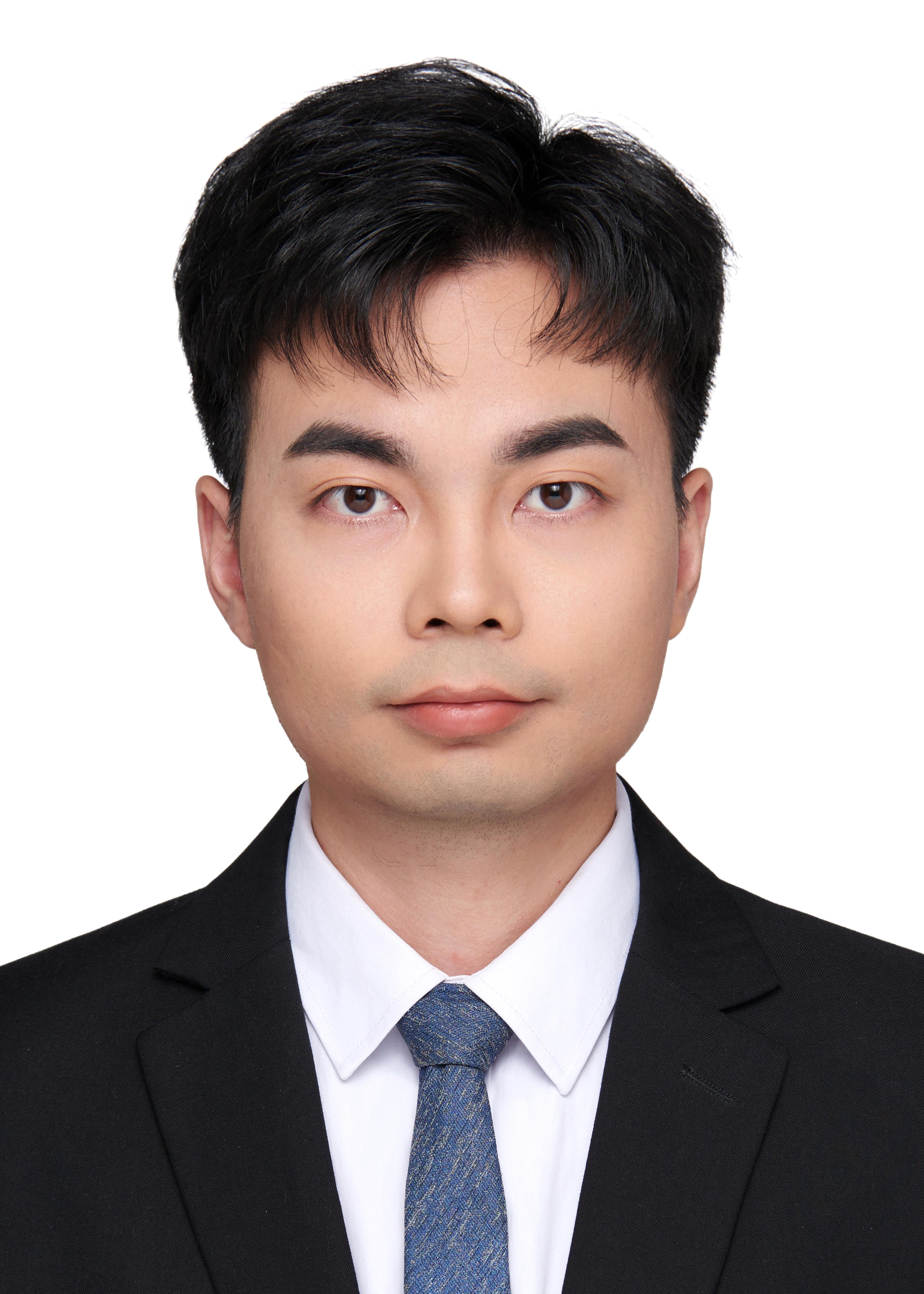}}] {Mengqi Huang} received the Ph.D. degree at the University of Science and Technology of China, in 2025. He has over ten publications that appeared in top-tier conferences, including CVPR, AAAI, and ACM MM. He was a recipient of the best student paper award in ACM Multimedia 2022. His research interests include image generation and deep generative models.
\end{IEEEbiography}

\begin{IEEEbiography}
[{\includegraphics[width=1in,height=1.25in,clip,keepaspectratio]{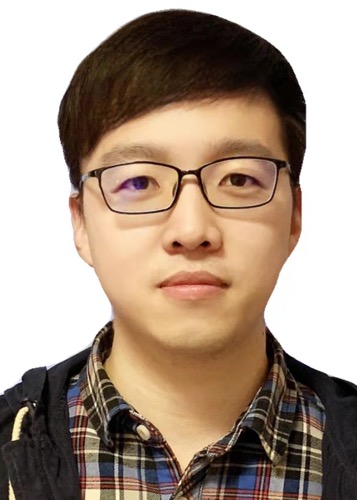}}]
{Fei Ding} received the M.S. degree in computer application technology from Renmin University of China, Beijing, in 2021. He is currently an Artificial Intelligence Researcher at Bytedance Inc.. His research interests are in the fields of computer vision, diffusion models, and multimodal models.
\end{IEEEbiography}

\begin{IEEEbiography}
[{\includegraphics[width=1in,height=1.25in,clip,keepaspectratio]{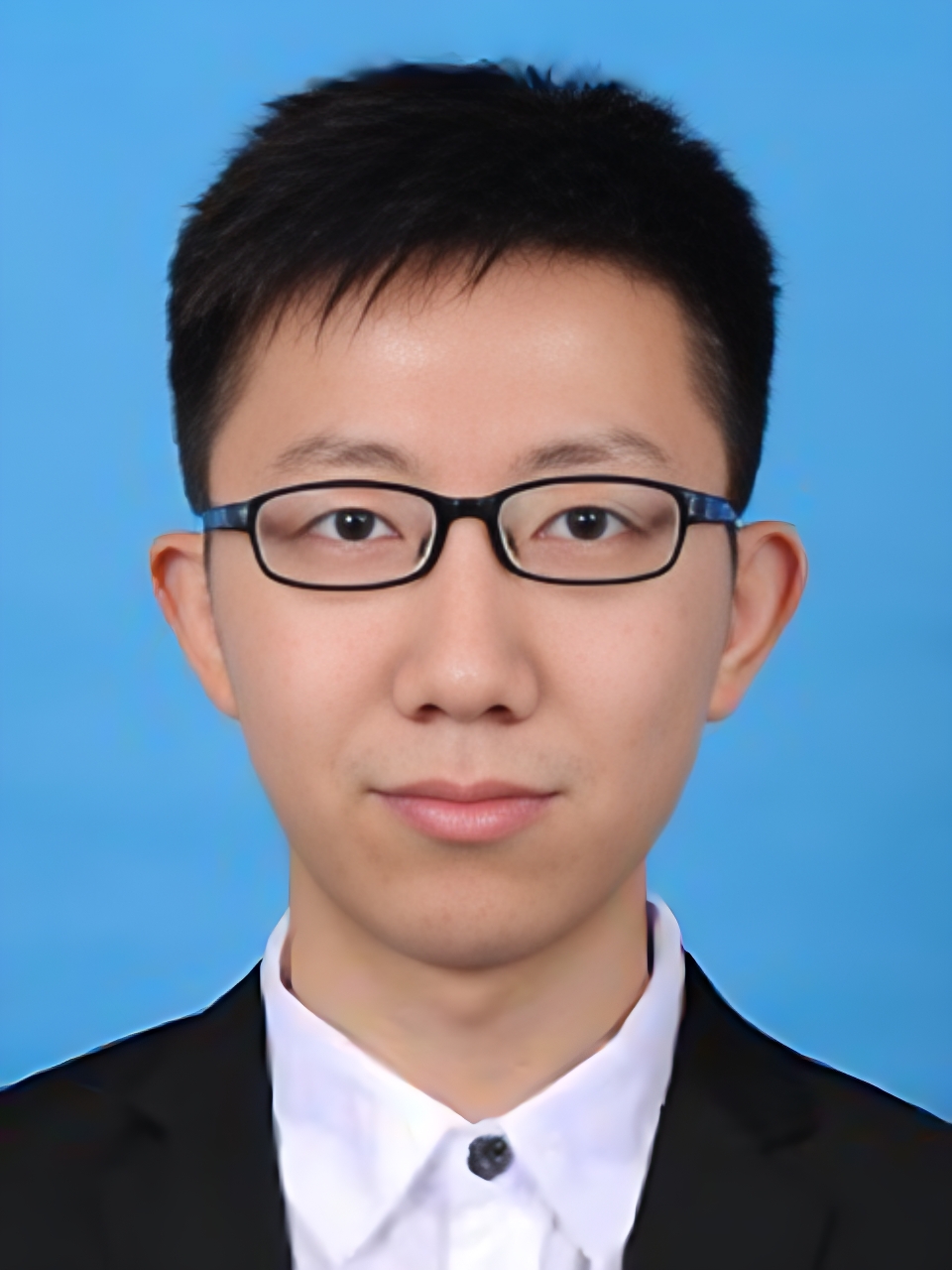}}]
{Mingcong Liu} received the M.S. degree in optical engineering from Beijing Institute of Technology in 2019. He is currently a Research Scientist at ByteDance Inc. Prior to this, he was a Research Engineer at Y-tech, Kuaishou Technology. His research interests include generative modeling, unsupervised learning, and image enhancement.
\end{IEEEbiography}

\begin{IEEEbiography}
[{\includegraphics[width=1in,height=1.25in,clip,keepaspectratio]{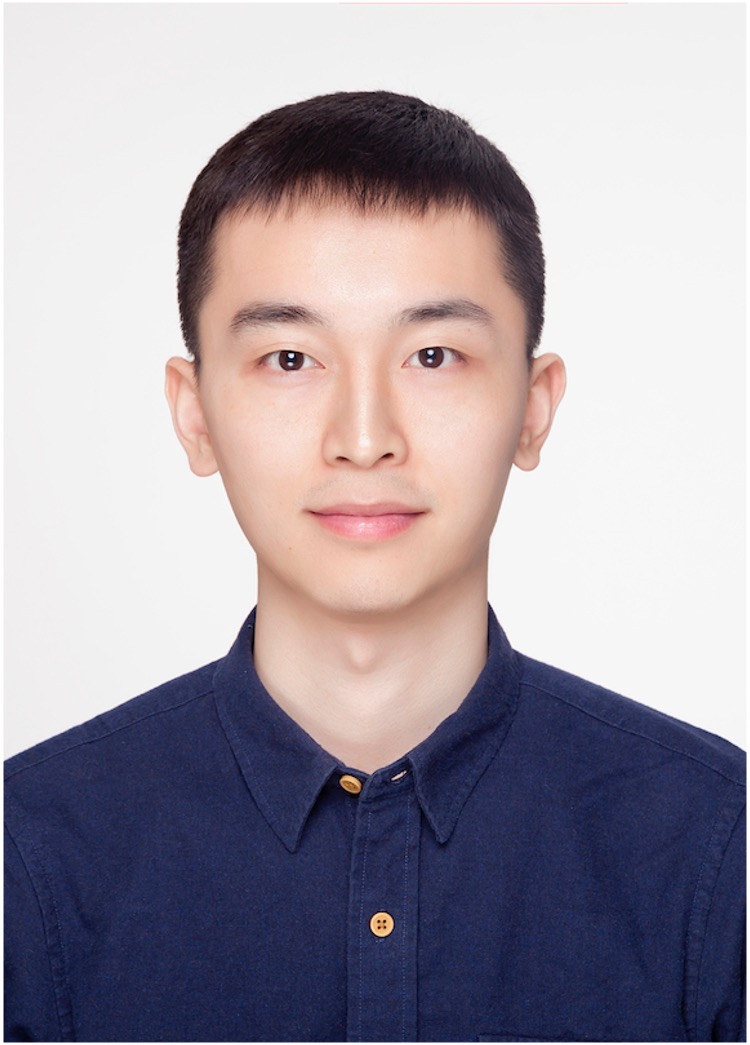}}]
{Qian He} obtained the M.S. degree from the Institute of Remote Sensing Applications of the Chinese Academy of Sciences in 2012. He joined Bytedance in 2017, and has been engaged in computer vision research ever since and has published more than 15 related papers, including CVPR, AAAI and ICCV. His research fields include image generation and editing, video generation and editing, and AIGC applications.
\end{IEEEbiography}

\begin{IEEEbiography}[{\includegraphics[width=1in,height=1.25in,clip]{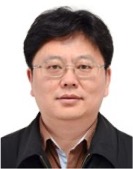}}]
{Yongdong Zhang} (M'08–SM'13-F'24) received the Ph.D. degree in electronic engineering from Tianjin University, Tianjin, China, in 2002. He is currently a Professor with the School of Information Science and Technology, University of Science and Technology of China. His current research interests are in the fields of multimedia content analysis and understanding, multimedia content security, video encoding, and streaming media technology. He has authored over 200 refereed journal and conference papers, accumulating more than 29,000 citations on Google Scholar. He was a recipient of the best paper awards in PCM 2013, ICIMCS 2013, ICME 2010, the best student paper award in ACM Multimedia 2022 and the Best Paper Candidate in ICME 2011. He serves as an Editorial Board Member of the \textit{Multimedia Systems Journal} and the \textit{IEEE Transactions on Multimedia}. He is a fellow of the IEEE.
\end{IEEEbiography}




\vfill

\end{document}